\documentclass[10pt,twocolumn,letterpaper]{article}

\usepackage{cvpr}              % To produce the CAMERA-READY version
\usepackage{amsmath}
\usepackage{graphicx}
\usepackage{colortbl}
\usepackage{tabularx}
\usepackage{booktabs}
\usepackage{amssymb} % For checkmarks
\usepackage{multirow}
\usepackage{arydshln}
\usepackage{tikz}
\usepackage{ulem}
\usepackage{multirow}
\usepackage[utf8]{inputenc}
\usepackage{caption}
\usepackage{makecell} % 在导言部分添加此包

\usepackage{float}
\newcommand{\cmark}{\checkmark} % Define checkmark symbol
\newcommand{\xmark}{\text{\sffamily X}} % Define x-mark symbol

\definecolor{cvprblue}{rgb}{0.21,0.49,0.74}
\usepackage[pagebackref,breaklinks,colorlinks,allcolors=cvprblue]{hyperref}

\def\paperID{7966} % *** Enter the Paper ID here
\def\confName{CVPR}
\def\confYear{2025}

\title{OmniStyle: Filtering High Quality Style Transfer Data at Scale}

\author{
    \makebox[\textwidth]{ % 让整行占满 textwidth，并居中
        Ye Wang$^{1}$ \quad Ruiqi Liu$^{1}$ \quad Jiang Lin$^{2}$ \quad Fei Liu$^{3}$ \quad Zili Yi$^{2}$ \quad Yilin Wang$^{4}$\footnotemark[1] \quad Rui Ma$^{1,5}$\thanks{Corresponding authors.}
    } \\
    \makebox[\textwidth]{{ $^1$Jilin University} \quad { $^2$Nanjing University} \quad  { $^3$ByteDance} \quad {$^4$Adobe}} \\
    \makebox[\textwidth]{{$^5$Engineering Research Center of Knowledge-Driven Human-Machine Intelligence, MOE, China}} \\
    \makebox[\textwidth][c]{Project page:} \\
    \makebox[\textwidth][c]{\url{https://wangyephd.github.io/projects/cvpr25_omnistyle.html}} \\
}

\begin{document}

\twocolumn[{%
\renewcommand\twocolumn[1][]{#1}%
\maketitle
\vspace*{-12pt}
\begin{center}
\centering
\includegraphics [width=1\linewidth]{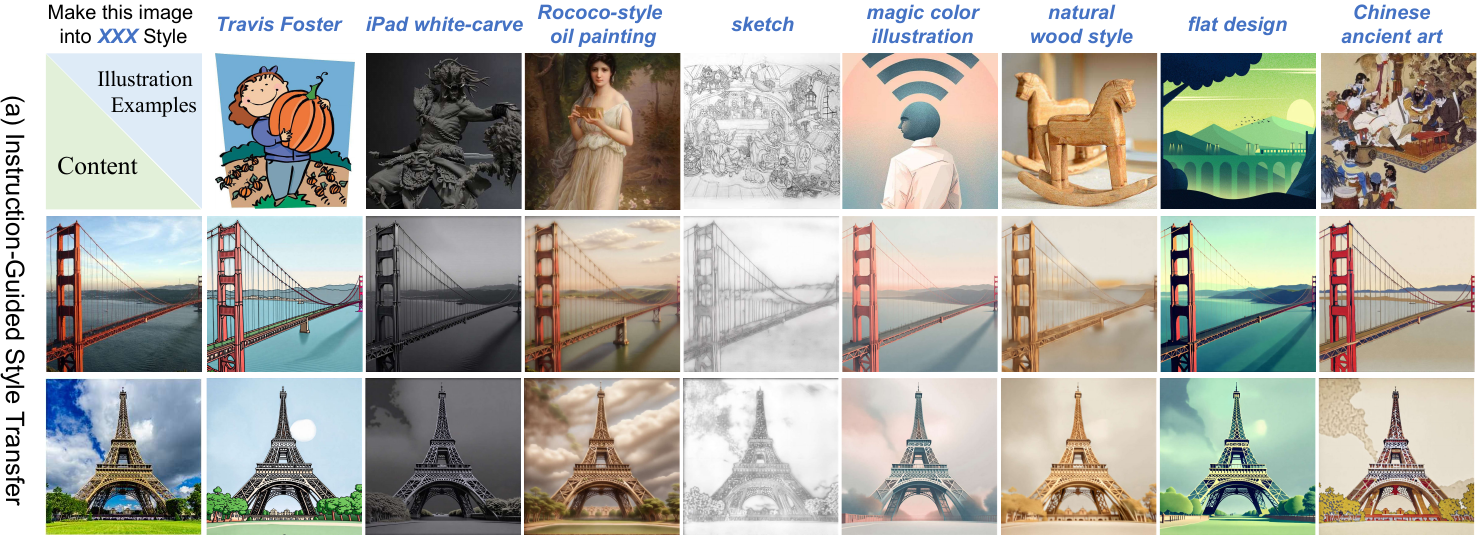}
% Add dashed line
\vspace{-10pt}
\begin{tikzpicture}
    \draw[dashed] (0,0) -- (\linewidth,0);
\end{tikzpicture}
\includegraphics [width=1\linewidth]{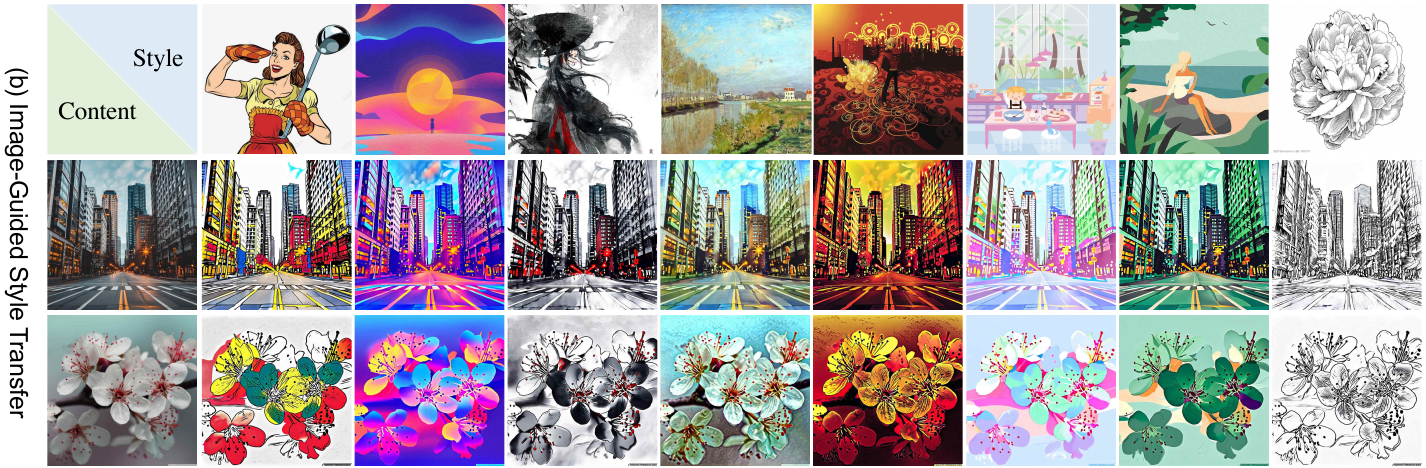}
\captionof{figure}{OmniStyle enables high-quality (a) instruction-guided style transfer and (b) reference image-guided style transfer, covering a diverse range of styles, including but not limited to comics, vector art, oil painting, sketch, and Chinese ancient art. Note that in (a), a style image of the style descriptions is provided for illustration, and our method only takes a text instruction and a content image as input. In (b), results are generated in a traditional manner of style transfer, in which the model takes both the content and style images as input. }
\label{fig:teaser}
\end{center}
}]

\renewcommand{\thefootnote}{\fnsymbol{footnote}}
\footnotetext[1]{Corresponding authors.}
\renewcommand{\thefootnote}{\arabic{footnote}} % 恢复正常编号格式

\begin{abstract}

In this paper, we introduce OmniStyle-1M, a large-scale paired style transfer dataset comprising over one million content-style-stylized image triplets across 1,000 diverse style categories, each enhanced with textual descriptions and instruction prompts. We show that OmniStyle-1M can not only enable efficient and scalable of style transfer models through supervised training but also facilitate precise control over target stylization. Especially, to ensure the quality of the dataset, we introduce OmniFilter, a comprehensive style transfer quality assessment framework, which filters high-quality triplets based on content preservation, style consistency, and aesthetic appeal. Building upon this foundation, we propose OmniStyle, a framework based on the Diffusion Transformer (DiT) architecture designed for high-quality and efficient style transfer. This framework supports both instruction-guided and image-guided style transfer, generating high resolution outputs with exceptional detail. Extensive qualitative and quantitative evaluations demonstrate OmniStyle's superior performance compared to existing approaches, highlighting its efficiency and versatility. OmniStyle-1M and its accompanying methodologies provide a significant contribution to advancing high-quality style transfer, offering a valuable resource for the research community.

\end{abstract}

\section{Introduction}
\label{sec:intro}
Style transfer \cite{gatys2016image}, the process of altering the artistic style of an image while preserving its content, has gained significant attention due to its creative potential across various domains, including digital art, advertising and fashion design. This technique has evolved rapidly, progressing  from early optimization-based methods \cite{gatys2016image,gatys2017controlling,kolkin2019style} to recent diffusion-based approaches \cite{wang2024instantstyle,wang2024instantstyleplus,xing2024csgo,gao2024styleshot,sohn2023styledrop}. 

While existing methods have shown promising results in transferring styles from reference images to content images, limitations remain in handling truly arbitrary styles. For instance, methods such as \cite{gatys2016image,johnson2016perceptual,huang2017arbitrary,kolkin2019style,li2017universal} focus on styles found in traditional artworks, such as oil painting, watercolor and sketches, while more recent methods \cite{zhang2023inversion, zhang2023prospect, ruiz2023dreambooth, sohn2023styledrop, luan2017deep} expand style transfer to include diverse styles like photography, cartoons, and illustrations. However, these methods still lack the generalization capability to handle truly arbitrary styles. Additionally, since many of these approaches rely on unsupervised training, they often lack control over the stylization process, e.g., unpredictable artifacts and distortion in stylized images. Moreover, due to the lack of large-scale, high-quality paired datasets, many methods are forced to adopt non-end-to-end architectures, such as optimization-based approaches \cite{gatys2016image, gatys2017controlling, kolkin2019style} and recent inversion-based techniques \cite{zhang2023inversion, zhang2023prospect, ruiz2023dreambooth, sohn2023styledrop}. These methods are computationally intensive, often requiring hundreds or even thousands of iterations to capture a specific style, or relying on inverting the content image and transferring style features within intermediate diffusion U-net blocks. These challenges substantially limit the efficiency and practicality of current methods for real-world applications. 

To address these challenges, we first introduce OmniStyle-1M, a large scale and high quality dataset which contains one million content-style-stylized image triplets across 1,000 distinct style categories, each enriched with detailed textual descriptions and instructional prompts. Specifically, we first generate content images in 20 categories using FLUX \cite{flux} and then select 1,000 style images from diverse categories within the Style30K dataset \cite{li2024styletokenizer}. Then, we employ six state-of-the-art (SOTA) style transfer models to produce stylized results, ensuring a broad range of data sources and stylized results. Further details on the dataset are provided in Section \ref{dataset}.

A paired dataset provides well-defined target styles for each content image, enabling models to generate stylized images with greater consistency and predictability. To control style transfer outputs, we introduce OmniFilter, a multi-dimensional quality assessment and filtering framework based on CLIP \cite{radford2021learning} and a Multimodal Large Language Model (MLLM) \cite{chen2023internvl}, tailored for customizing the OmniStyle-1M dataset. OmniFilter comprehensively evaluates image quality  across three key dimensions: content preservation, ensuring that the original structure and semantics of the content image are retained; style consistency, guaranteeing alignment between the transferred and reference styles; and aesthetic appeal, ensuring that the output is visually engaging.

Leveraging the high-quality, diverse data filtered by OmniFilter, we present OmniStyle, an end-to-end framework for high-quality and efficient style transfer based on the Diffusion Transformer (DiT) architecture \cite{flux, xiao2024omnigen,zhou2024transfusion,peebles2023scalable}. OmniStyle supports both instruction-guided and image-guided style transfer tasks, generating rich, detailed results at high resolution. Experimental results demonstrate that our model achieves superior performance compared to existing style transfer methods. Our contributions are follows:

\begin{itemize}
    \item We propose \textbf{OmniStyle-1M}, the first million-level paired style transfer dataset, containing 1,000 style categories, along with a rich collection of textual and instruction prompt data. Such data paves a way for research community on enhancing supervised model training, controlling predictable stylized results and handling more diverse style references.

    \item We propose \textbf{OmniFilter}, a specialized stylization-driven framework for assessing and filtering high-quality stylized images. OmniFilter evaluates images based on content preservation, style consistency, and aesthetic appeal, allowing models to learn from explicitly filtered high-quality examples for effective style transfer.

    \item We propose \textbf{OmniStyle}, an efficient, flexible, simple feed-forward style transfer framework, which can serve as a strong baseline for future style transfer research and is adaptable for real-world applications through further model distillation.
\end{itemize}

\begin{table*}[]
\caption{Comparison of style-related datasets. \(\xmark\) denotes `not included', $\scriptstyle\checkmark$ denotes `few included' and \(\checkmark\) denotes `included'.}
\label{tab:dataset}
\resizebox{\textwidth}{!}{%
\begin{tabular}{cccccccc}
\hline
\multirow{2}{*}{Datasets} & \multirow{2}{*}{Main Task} & \multicolumn{3}{c}{Triplet Components}  & \multirow{2}{*}{Style Category Number} & \multirow{2}{*}{Prompt Category} & \multirow{2}{*}{Triplet Number} \\
                          &                                & content & style & stylized &                                        &                                  &                              \\ \hline
EditBench \cite{wang2023imagen}                & Image Editing                                & $\scriptstyle\checkmark$        & \xmark      & $\scriptstyle\checkmark$         & ---                                    & Instruction                      & ---                              \\
MagicBrush \cite{Zhang2023MagicBrush}               & Image Editing                                & $\scriptstyle\checkmark$        & \xmark      & $\scriptstyle\checkmark$          & ---                                    & Instruction                      & ---                              \\
HQ-Edit \cite{hui2024hq}                  & Image Editing                                & $\scriptstyle\checkmark$        & \xmark      & $\scriptstyle\checkmark$          & ---                                    & Instruction                      & ---                              \\
UltraEdit \cite{zhao2024ultraedit}                  & Image Editing                                & $\scriptstyle\checkmark$        & \xmark      & $\scriptstyle\checkmark$          & ---                                    & Instruction                      & ---                              \\

InstructPix2Pix \cite{brooks2022instructpix2pix}           & Image Editing                                & $\scriptstyle\checkmark$        & \xmark      & $\scriptstyle\checkmark$          & ---                                    & Instruction                      & ---                              \\ \hline
Style30K \cite{li2024styletokenizer}                 & Style-Related                               &  \xmark        & \cmark      & \xmark          & 1,120                                  & Style Category                   & 30K (Only Style Images)                           \\
WikiArt \cite{saleh2015large}              & Style-Related                               &  \xmark        & \cmark      & \xmark          & 27                                  & Style Category                   & 57K (Only Style Images)                           \\
ArtBench \cite{liao2022artbench}                   & Style-Related                               &  \xmark        & \cmark      & \xmark          & 10                                  & Style Category                   & 60K (Only Style Images)                           \\ \hline
IMAGStyle \cite{xing2024csgo}                & Style-Related                               & \cmark         & \cmark      & \cmark          & 14                                     & Text, Image                      & 210K                          \\ \hdashline
OmniStyle-1M               & Style-Related                             & \cmark         & \cmark      & \cmark          & 1000                                   & Text, Image, Instruction           & 1M                            \\ \hline
\end{tabular}
}

\end{table*}

\section{Related Work}
\label{sec:related_work}

\noindent{\textbf{Style Transfer.}} 
Style transfer has evolved from early handcrafted feature-based methods \cite{zhang2013style,wang2004efficient} to optimization-driven approaches \cite{gatys2016image, gatys2017controlling, kolkin2019style}, followed by arbitrary style transfer models \cite{deng2020arbitrary,huang2017arbitrary,liao2017visual,zhang2022exact,li2017universal}, and more recently, diffusion-based methods \cite{wang2024instantstyle,wang2024instantstyleplus,chung2024style,sohn2023styledrop,zhang2023inversion,jeong2023training,xu2024freetuner,gao2024styleshot,xing2024csgo}, which achieve superior performance through tuning-based \cite{zhang2023inversion,zhang2023prospect,wang2023stylediffusion} and tuning-free \cite{wang2024instantstyle,wang2024instantstyleplus,xing2024csgo,gao2024styleshot,qi2024deadiff} strategies. However, these methods are often computationally expensive and time-consuming due to the need for extensive iterative optimization or the introduction of additional DDIM inversion, making them impractical for large-scale or real-time use. Additionally, many models lack flexibility, as they are typically designed for either image-guided or text-guided transfer, restricting their versatility in multi-task scenarios. In this paper, we address these limitations by introducing OmniStyle-1M, a high-quality, large-scale dataset that supports the training of an end-to-end, efficient style transfer framework capable of handling diverse style transfer tasks with improved flexibility.

% Existing models \cite{xing2024csgo,wang2024instantstyle,wang2024instantstyleplus} lack scalability, handling only a limited number of style categories. 
% Many methods still rely on inefficient, resource-intensive training processes, particularly in optimization-based \cite{gatys2016image,gatys2017controlling,kolkin2019style}  and diffusion-tuning approaches \cite{zhang2023inversion,zhang2023prospect,ruiz2023dreambooth,sohn2023styledrop}, which require extensive iterations and struggle to generalize to unseen styles. Additionally, most methods \cite{artflow2021,chung2024style, hong2023aespa,gatys2016image,gatys2017controlling,kolkin2019style} fail to achieve high-resolution outputs with sufficient detail, making them unsuitable for applications demanding fidelity. 

\noindent{\textbf{Style-Related Datasets.}} 
Table \ref{tab:dataset} compares various style-related image editing datasets. The first category represents image editing task datasets, such as UltraEdit \cite{zhao2024ultraedit}, which only include a small number of content-stylized pairings and limited textual instructions, making it inadequate for handling a wide variety of style categories. The second category refers to datasets that only include style images, such as Style30K \cite{li2024styletokenizer}, which covers 1,120 fine-grained style categories. However, due to the lack of paired data, this dataset can only facilitate style-guided text-to-image generation and cannot support style transfer tasks. IMAGStyle \cite{xing2024csgo} and our OmniStyle-1M are the only two datasets that contain triplet data. Compared to IMAGStyle, OmniStyle-1M has several advantages: 1) OmniStyle-1M contains 1,000 fine-grained style categories, far surpassing the 14 categories in IMAGStyle; 2) OmniStyle-1M is derived from six SOTA models, ensuring diverse data sources, whereas IMAGStyle relies solely on B-LoRA \cite{frenkel2024implicit} for data generation, limiting its diversity; 3) Our dataset consists of over 1 million triplet data points, five times the size of IMAGStyle. In summary, OmniStyle-1M offers a more comprehensive and diverse dataset compared to existing ones, making it more suitable for advancing style transfer research and applications.

\section{OmniStyle-1M}
\label{dataset}

\begin{figure*}[t]
    \centering
    \includegraphics[width=\textwidth]{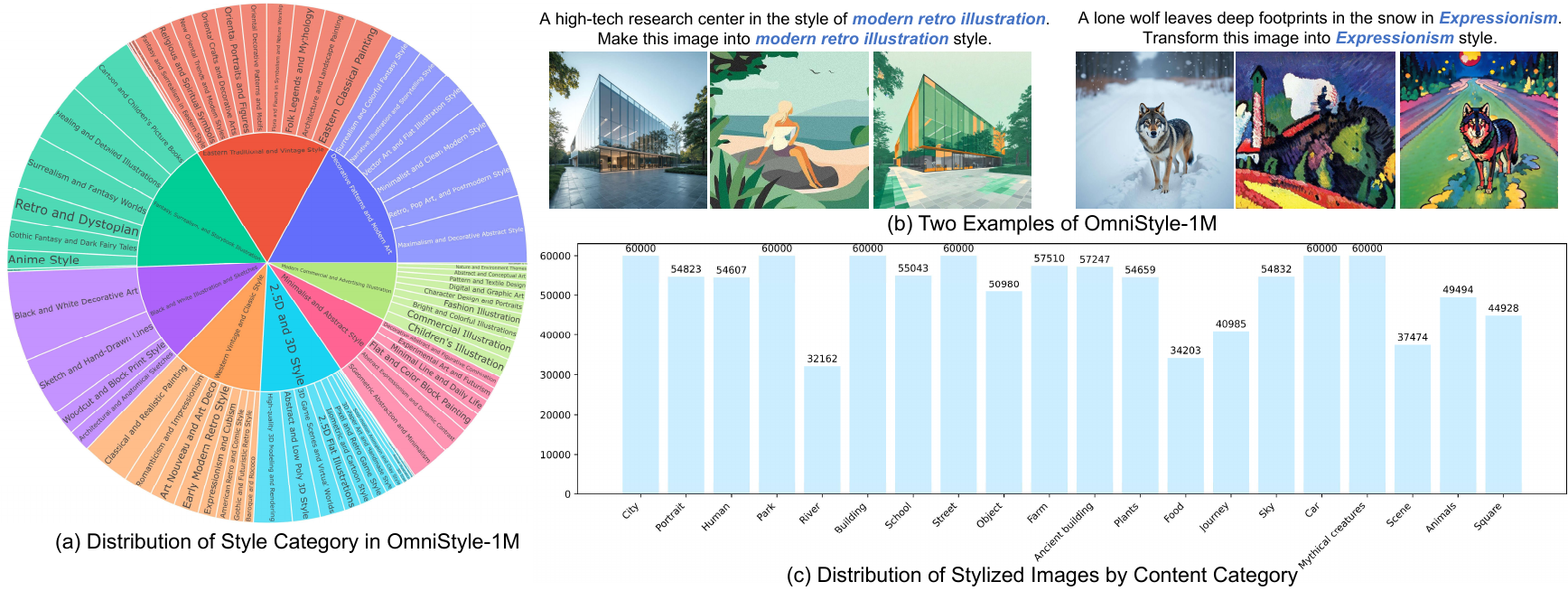}
    \caption{Overview of OmniStyle-1M. (a) The inner ring represents the eight primary categories, while the outer ring corresponds to specific fine-grained classifications, illustrating the extensive diversity of style categories within our dataset. (b) Two examples of triplets are shown, each includes a content image, a style image, a stylized output, a corresponding textual description, and an instructional prompt. (c) Distribution of stylized results across different content categories. }
    
    \label{fig:dataset}
\end{figure*}

We start with an overview of OmniStyle-1M in Section \ref{dataset_overview}, followed by a detailed explanation of the data curation process in Section \ref{Dataset_curation}. Next, we introduce OmniFilter, our MLLM-based data evaluation and filtering framework, in Section \ref{sec:mllm}.

\subsection{Dataset Overview}
\label{dataset_overview}

OmniStyle-1M represents the first large-scale content-style-stylized image triplets dataset created specifically for style transfer tasks. As depicted in Figure \ref{fig:dataset}.a, it features 1,000 distinct and fine-grained style categories, e.g., American comics, Eastern art, vector art, line drawings, sketches, cartoons, and intricate paintings. Additionally, as shown in Figure \ref{fig:dataset}.c, OmniStyle-1M spans 20 different content image categories, covering cityscapes, parks, humans, animals, plants, scenes, objects, food, and more. The dataset maintains a balanced distribution of stylized images across these content categories, effectively avoiding severe long-tail distribution issues. Figure \ref{fig:dataset}.b further illustrates two triplet examples, each comprising a content image, a style image, and a stylized result image, alongside two unique prompt texts and instructions. The following section details the dataset's curation process.

\subsection{Dataset Curation}
\label{Dataset_curation}
\begin{figure*}[t]
    \centering
    \includegraphics[width=\textwidth]{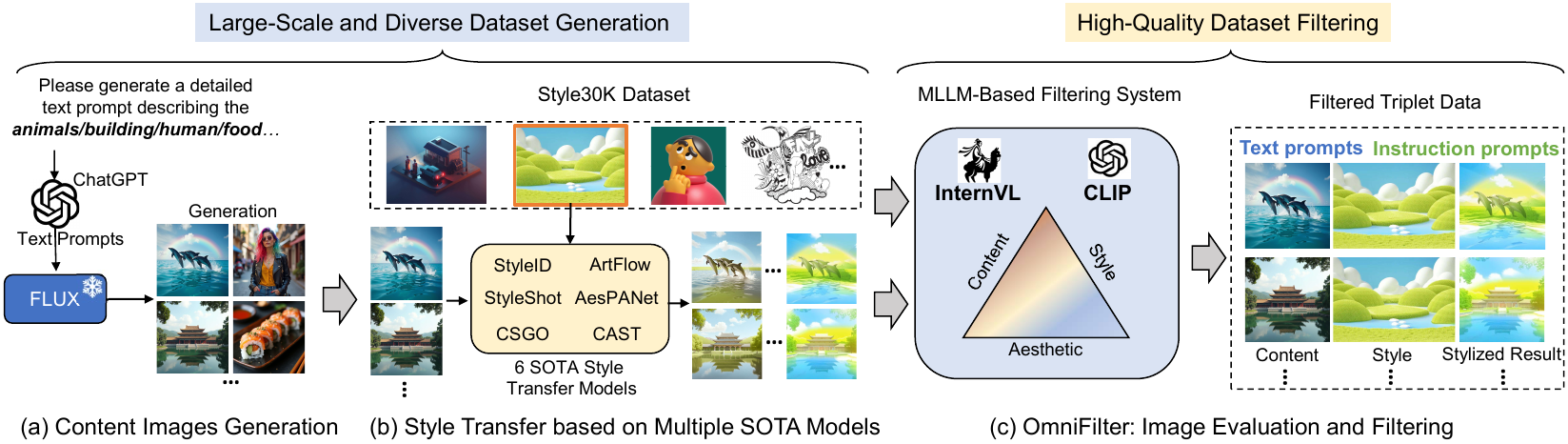}
    \caption{Overview of our dataset creation and filtering pipeline. (a) Content Image Generation: We utilize ChatGPT to automatically generate textual descriptions across 20 categories (e.g., animals, architecture, humans, food) and generate corresponding images using the FLUX model. (b) Style Transfer: Style images are randomly sampled from the Style30K dataset, and six SOTA style transfer models are applied to generate a large and diverse dataset of one million triplets. (c) OmniFilter: Stylized images are filtered based on content consistency, style preservation, and aesthetic appeal to ensure high-quality, visually cohesive results.}
    
    \label{fig:method}
\end{figure*}

Our dataset curation pipeline is illustrated on the left side of Figure \ref{fig:method}. It consists of two main stages: generation of content images and style transfer using multiple state-of-the-art models to obtain stylized images.

\noindent{\textbf{Content Image Generation}}. As shown in Figure \ref{fig:method}.a, we utilize FLUX \cite{flux} and ChatGPT \cite{openai2024chatgpt} to generate a diverse set of content images. We begin by defining 20 common content categories (see Figure \ref{fig:dataset}.c) and then use ChatGPT to automatically generate extensive textual prompts for each category. The template for querying follows the format ``Please generate a detailed text prompt describing [category],'' where ``category" is replaced by specific topics such as animals, architecture, humans, and so forth. We generate 100 prompts per category, resulting in a dataset of 2,000 prompts in total. These prompts are then input into FLUX \cite{flux} to produce high-quality content images with rich details without copyright issues. Ultimately, this process yields 2,000 unique and detailed images, each corresponding to its specific category. To ensure diversity and realism, prompts are designed to encompass a wide range of subtopics within each category. For example, the ``animals" category includes various animal types in different environments and poses, while the ``architecture" category covers modern, classical, and abstract structures. This approach ensures that the generated dataset is comprehensive and unbiased.

\noindent{\textbf{Stylized Image Generation}}.
Building on the generated high-quality content images, we move to the stylization phase, as illustrated in Figure \ref{fig:method}.b. Unlike IMAGStyle \cite{xing2024csgo}, which solely relies on B-LoRA \cite{frenkel2024implicit} for data generation, our pipeline integrates six advanced style transfer models: StyleID \cite{chung2024style}, ArtFlow \cite{artflow2021}, StyleShot \cite{gao2024styleshot}, AesPANet \cite{hong2023aespa}, CSGO \cite{xing2024csgo}, and CAST \cite{zhang2020cast}. This diverse selection enhances the variety and richness of the generated data. We curate a set of 1,000 style images from various categories within the Style30K \cite{li2024styletokenizer} dataset, ensuring coverage of a broad range of styles. For each content image, we randomly select 100 style images from this set and apply style transfer using each of the six models individually. This approach ensures that every content image undergoes six unique stylizations, spanning a wide variety of styles and addressing the limitations of single-source data generation. As a result, we produce OmniStyle-1M, a dataset containing one million stylized image triplets.

\subsection{OmniFilter }
\label{sec:mllm}
To further control the style transfer output, we design OmniFilter to identify and extract high-quality stylized triplets from the OmniStyle-1M dataset. As shown in Figure \ref{fig:method}.c, by assessing each triplet based on three key criteria: content preservation, style consistency, and aesthetic appeal. OmniFilter ensures that only the highest-quality triplets are selected. This process enhances the dataset's quality for downstream style transfer tasks.

\noindent{\textbf{Content Preservation Evaluation.}} We evaluate content preservation between the stylized and content images from two main perspectives: semantic consistency and structural integrity. Semantic consistency measures the alignment of high-level semantic content between the stylized image and the content image, ensuring that key subjects and important objects remain identifiable after stylization. Structural integrity assesses the preservation of spatial layout and geometric features, determining whether the stylization maintains the essential structure of the image, thereby preserving the overall visual composition.

For semantic consistency, we use the CLIP \cite{radford2021learning} image-text similarity score, which compares the stylized image to the content image’s caption. The caption provides a detailed description of key semantic elements, and the similarity score is treated as the measure of semantic consistency. Since CLIP \cite{radford2021learning} is not suitable for evaluating structural preservation, we instead use the self-supervised model DINOv2 \cite{oquab2023dinov2}. By extracting embeddings from both the stylized and content images using DINOv2 \cite{oquab2023dinov2}, we compute the similarity between these embeddings to obtain the structural preservation score. Finally, we combine the semantic consistency score, $S_{\text{semantic}}$, and the structural preservation score, $S_{\text{structural}}$, to calculate the overall content preservation score $C_{\text{score}}$:

\begin{equation} C_{\text{score}} = \alpha \cdot S_{\text{semantic}} + (1 - \alpha) \cdot S_{\text{structural}}, \end{equation}
where $\alpha$ is a weighting factor that controls the balance between the two scores, and $\alpha$ is set to 0.5.

% \begin{figure*}[t]
%     \centering
%     \includegraphics[width=\linewidth]{assets_downsample/content_style_score.pdf}
%     \caption{TODO}
    
%     \label{fig:dataset}
% \end{figure*}

\noindent{\textbf{Style Consistency Evaluation.}}  
Evaluating style consistency is inherently challenging due to its subjective and abstract nature. Different individuals may perceive the same style in varying ways, and styles often comprise multiple visual components that are difficult to quantify, making objective assessments more complex. While existing metrics like style loss can measure style similarity in specific scenarios, they often lack the robustness required to capture the full range and intricacies of different styles. To address this challenge, we employ a self-supervised learning approach to assess style consistency scores. Specifically, we use the publicly available Style30K \cite{li2024styletokenizer} dataset, where images from the same style are treated as positive sample pairs and those from different styles as negative pairs. Through contrastive learning, we fine-tune the CLIP \cite{radford2021learning} image encoder, which allows images of the same style to group together in the feature space, while images from different styles are effectively separated. This fine-tuned CLIP \cite{radford2021learning} encoder can then compute a style consistency score $S_{score}$ by evaluating the similarity between the embeddings of two styles.

\noindent{\textbf{Aesthetic Appeal Evaluation.}}
Aesthetic appeal is a vital aspect in the evaluation of stylized images, as it directly influences users' visual experience and emotional response. However, many existing evaluation methods \cite{xing2024csgo, gao2024styleshot} fail to adequately address this dimension. Recent work by ViPer \cite{salehi2024viper} emphasizes the significant role of visual attributes in shaping individual aesthetic preferences. Builing upon it, we introduce a novel approach for aesthetic evaluation that leverages the InternVL2 \cite{chen2023internvl} model based on multimodal visual attribute features.

We first define 40 key visual attributes essential for assessing aesthetic quality, such as composition, balance, color harmony, lighting, contrast, and saturation, etc. Using these attributes, we design query prompts to enable InternVL2 \cite{chen2023internvl} to describe the visual attributes of each image, generating attribute-based textual captions. We use two key datasets for training and evaluation: the AVA dataset \cite{murray2012ava}, which contains natural images with aesthetic scores, and the BAID dataset \cite{Yi_2023_CVPR}, which focuses on artistic images rated for aesthetic appeal. After extracting the visual attribute descriptions for each image, we train the model to predict aesthetic scores based on both the image features and their corresponding visual attribute captions, utilizing ground-truth aesthetic scores to minimize MSE loss.

Specifically, we employ InternVL-G \cite{chen2023internvl} to extract the image features and their associated attribute textual description features. These features are then concatenated and passed through an MLP projection layer to perform the aesthetic appeal scoring $A_{score}$. The training process begins with the AVA dataset \cite{murray2012ava}, followed by fine-tuning on the BAID dataset \cite{Yi_2023_CVPR}. This fine-tuning step bridges the domain gap between natural and stylized images, enabling the model to better assess aesthetic appeal in stylized images.

\noindent{\textbf{Data Filtering.}} For each content-style image pair, we generated six stylized outputs using six different state-of-the-art style transfer models. For each output, we calculated scores across content, style, and aesthetic dimensions, and combined these scores to get a total score for each output. The combined score is calculated as follows: 
\begin{equation} 
\textit{Score} = a \cdot C_{\text{score}} + b \cdot S_{\text{score}} + c \cdot A_{\text{score}},
\end{equation}  
where $a=0.2$, \(b=0.6\), and \(c=0.2\). The output with the highest total score among the six was selected as the final sample.
This filtering yielded a high-quality subsets: OmniStyle-150K with 150K triplets across 1,000 categories. These high-quality data subsets provide robust support for our model training.

\section{OmniStyle}
\label{sec:method}

\subsection{Architecture}
\label{omni_archi}

As shown in Figure \ref{fig:architecture}, OmniStyle is a style transfer framework based on the FLUX-dev \cite{flux} model, capable of supporting high-quality style transfer tasks guided by either instructions or reference images. 
Specifically, for reference-guided style transfer, we employ VAE to extract continuous visual features from both the style and content images. These features are then spatially concatenated with noisy latents and text tokens before being fed into MM-DiT. Additionally, different positional encodings are applied separately to the style and content features. 
For instruction-guided style transfer, the content feature sequence is treated as text tokens and spatially concatenated with the extracted text tokens and noisy latents. To distinguish content features, special tags $[img]$ and $[/img]$ are assigned within the sequence. 
Similarly, for reference-based style transfer, the image sequences are marked with $[img1][/img1]$ and $[img2][/img2]$ to  distinguishes style and content images.

\subsection{Implementation Details}
\label{omni_implement}
\begin{figure}
\centering
\includegraphics[width=0.85\linewidth]{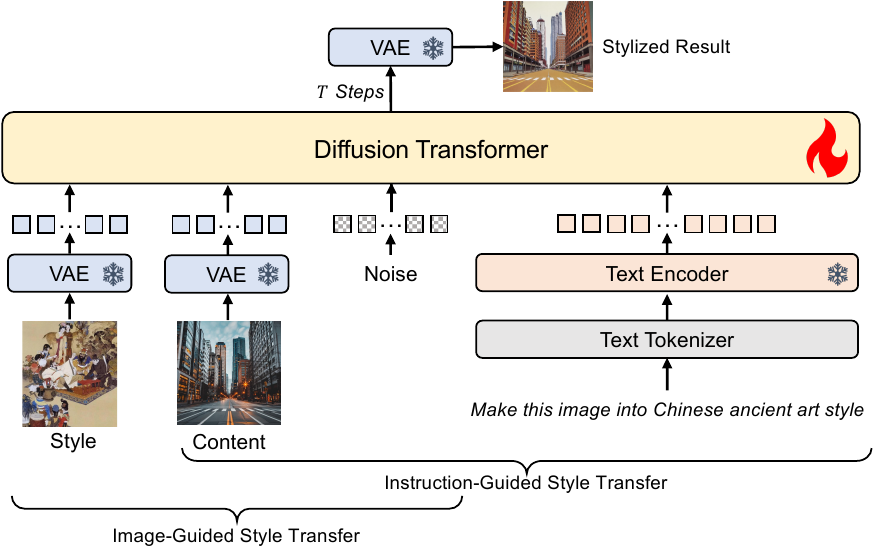}
\caption{The architecture of OmniStyle. }
\label{fig:architecture}
\vspace{-1.7em} 
\end{figure}

Our model is trained on 8 × NVIDIA H100 GPUs, with a batch size of 1 per GPU and a learning rate of 1e-4.  As shown in Figure  \ref{fig:architecture}, only the diffusion transformer is finetuned with style transfer dataset while all other components are kept as frozen. We apply random cropping and horizontal flipping to the input style image to improve the robustness for style learning.

\label{sec:Implementation}

\section{Experiments}
\label{sec:experiments}

\noindent{\textbf{Evaluation Benchmark.}}  
For a comprehensive evaluation, we develop a unified benchmark for image-guided and instruction-guided style transfer tasks, which includes 20 content images and 100 style images, resulting in a total of 2000 stylized images. For the image-guided task, the 100 style images were directly applied to generate stylized results. In the instruction-guided style transfer task, the style images were replaced with corresponding instruction prompts to guide the model to perform stylization. We hope this benchmark will serve as a standard for evaluating future style transfer methods and advancing the field.

\noindent{\textbf{Evaluation Metrics}.}
We use OmniFilter as a evaluation tool to assess the quality of stylization produced by different methods. Specifically, we employ three key metrics: content preservation score, style consistency score, and aesthetic appeal score, to evaluate the results. Additionally, style loss is utilized to help assess style consistency.

\noindent{\textbf{Comparison Methods}.}
\begin{itemize}
    \item Instruction-guided style transfer: we evaluated our approach against four instruction-based image editing methods, including InstructPix2Pix \cite{brooks2022instructpix2pix}, HQ-Edit \cite{hui2024hq}, UltraEdit \cite{zhao2024ultraedit}, and OmniGen \cite{xiao2024omnigen}, as well as a text-guided style transfer method, DiffStyler \cite{huang2024diffstyler}.

    \item Image-guided style transfer: we compared our approach with seven SOTA methods, including ArtBank\cite{zhang2024artbank}, StyleID \cite{chung2024style}, ArtFlow \cite{artflow2021}, StyleShot \cite{gao2024styleshot}, AesPANet \cite{hong2023aespa}, CSGO \cite{xing2024csgo}, and CAST \cite{zhang2020cast}.
\end{itemize}

\begin{table}[]
\caption{Quantitative comparison with respect to the SOTA methods on instruction-guided style transfer.}
\label{tab:style_quantitative_txt}
\resizebox{\linewidth}{!}{%
\setlength{\tabcolsep}{2.5pt}
\begin{tabular}{ccccc}
\hline
Method    & \makecell{Content \\ Preservation $\uparrow$} & \makecell{Style \\ Consistency $\uparrow$} & \makecell{Aesthetic \\ Appeal $\uparrow$} & \makecell{Style \\  \quad Loss $\downarrow$}  \\ \hline
InstructPix2Pix \cite{brooks2022instructpix2pix}   &  0.4848                          &  0.4965                      & 5.7446  & 0.3829                       \\
HQ-Edit \cite{hui2024hq} &  0.4508       & 0.2254                       &  \textbf{5.7777}      & 0.3108                  \\
UltraEdit \cite{zhao2024ultraedit}     &  0.4906                          &  0.5087                      & 5.6351      & 0.3432                   \\ 
OmniGen \cite{zhao2024ultraedit}     &  0.4918                          &  0.5487                      &  5.6717         & 0.6272               \\ 
DiffStyler \cite{huang2024diffstyler}     &  0.4816                          &  0.5127                      & 5.4551      & 0.4256             \\ \hline 
OmniStyle &  \textbf{0.5128}                          &  \textbf{0.6441}                      & 5.7512            & \textbf{0.2873}             \\ \hline

\end{tabular}
}
\vspace{-5pt}
\end{table}

\begin{figure*}[t]
    \centering
    \includegraphics[width=0.94\textwidth]{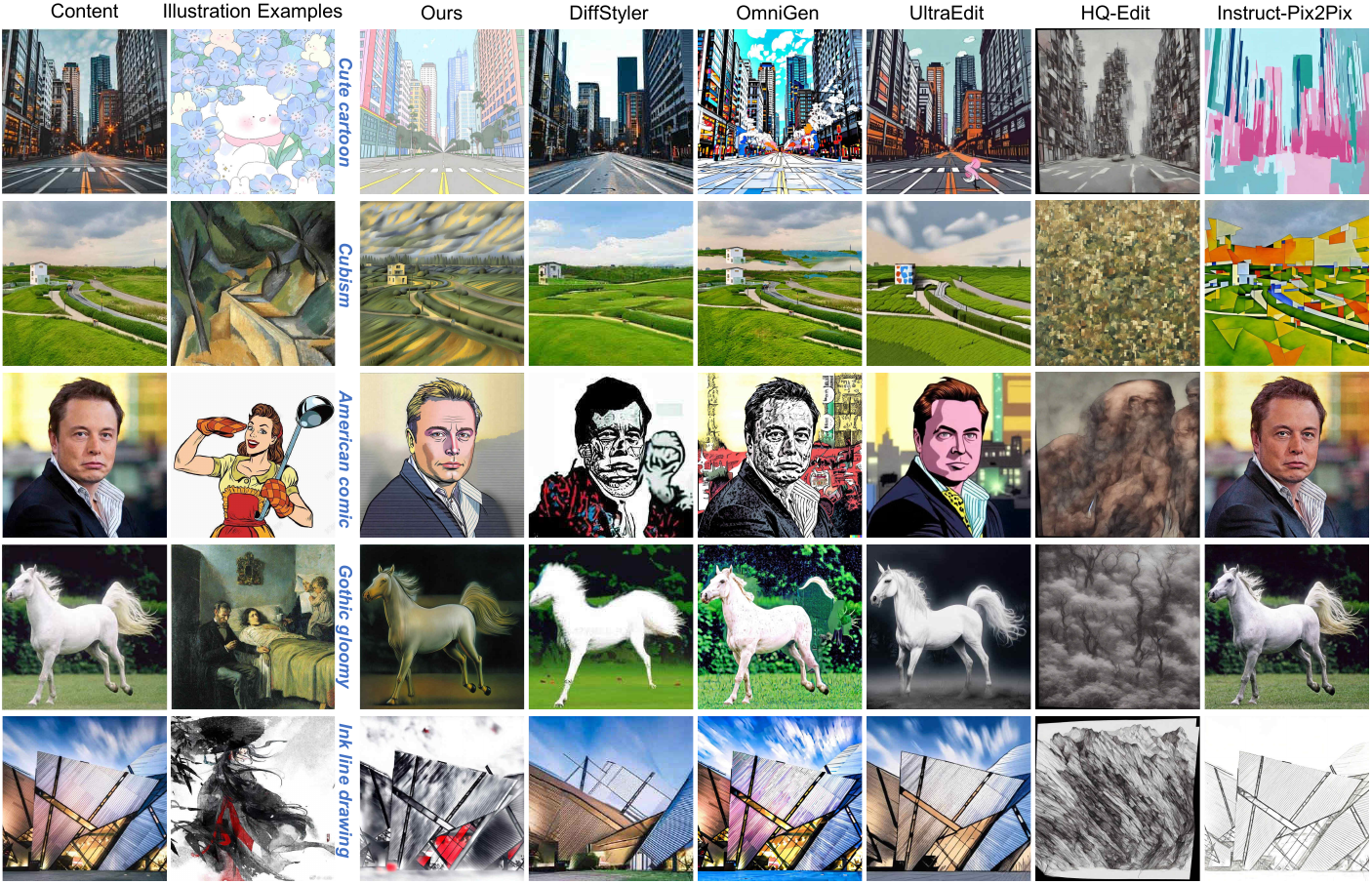}
    \caption{Qualitative comparison with other state-of-the-art methods for the instruction-guided style transfer task. For clarity, the style images and style categories are placed on the right side of the content images for reference.}
    
    \label{fig:instruction_compare}
    % \vspace{-1.5em} 
\end{figure*}

\subsection{Quantitative Experiments}

\noindent{\textbf{Instruction-Guided Style Transfer}}.  As shown in Table \ref{tab:style_quantitative_txt}, when compared to SOTA instruction-guided image editing methods, such as InstructPix2Pix, HQ-Edit, UltraEdit, and OmniGen, OmniStyle achieves the highest content preservation score (0.5128) and style consistency (0.6441), indicating its stronger ability to maintain content and style consistently. It also outperforms these methods in style loss (0.2873), demonstrating better style fidelity. In terms of aesthetic appeal, OmniStyle (5.7512) is competitive with HQ-Edit (5.7777), showcasing its ability to generate high-quality, visually appealing results. In comparison with DiffStyler, a leading text-guided style transfer method, OmniStyle excels in content preservation (0.5128 vs. 0.4816), style consistency (0.6441 vs. 0.5127), and style loss (0.2873 vs. 0.4256), while also achieving a higher aesthetic appeal (5.7512 vs. 5.4551). These results highlight OmniStyle's effectiveness in the instruction-guided style transfer task, surpassing existing methods.

\vspace{1cm}

\noindent{\textbf{Image-Guided Style Transfer}}. As shown in Table \ref{tab:style_quantitative_img}, OmniStyle outperforms existing SOTA methods for the image-guided style transfer task, particularly in terms of style consistency, aesthetic appeal, and style loss. It achieves the highest style consistency score (0.7483) and aesthetic appeal score (5.7913), surpassing methods such as StyleShot (0.7347, 5.7818) and CSGO (0.7251, 5.7712). Furthermore, OmniStyle leads in style loss with the lowest score (0.1086), demonstrating superior style fidelity. 

% style transfer

\begin{table}[]
\caption{Quantitative comparison with respect to the SOTA methods for the image-guided style transfer.}
\label{tab:style_quantitative_img}
\resizebox{\linewidth}{!}{
\setlength{\tabcolsep}{2.5pt}
\begin{tabular}{ccccc}
\hline
Method    & \makecell{Content \\ Preservation $\uparrow$} & \makecell{Style \\ Consistency $\uparrow$} & \makecell{Aesthetic \\ Appeal $\uparrow$} & \makecell{Style \\  \quad Loss $\downarrow$}\\ \hline
ArtFlow \cite{artflow2021}   &  0.5184                          &  0.6904                      &  5.1220       &  0.1289               \\
AesPANet \cite{hong2023aespa} &  0.5665                          &   0.6818                     &  5.2117         & 0.1128               \\
CAST \cite{zhang2020cast}      &  \textbf{0.6165}                          &  0.6770                      & 5.4561      & 0.1428                   \\
StyleID \cite{chung2024style}   & 0.5121                           & 0.6564                       &   5.3531       & 0.1315               \\
CSGO \cite{xing2024csgo}     &  0.5067                          &     0.7251                   &   5.7712       & 0.3727               \\
StyleShot \cite{gao2024styleshot} &  0.5410                          &  0.7347                      &  5.7818     & 0.1500                  \\ 
ArtBank \cite{zhang2024artbank}     &  0.3490                          &  0.7163                      & 5.4980      & 0.1208             \\ \hline 

OmniStyle &  0.5450                       & \textbf{0.7483}                           &  \textbf{5.7913 }     &  \textbf{0.1086}           \\ \hline
\end{tabular}
}
\vspace{-1.5em} 
\end{table}

\begin{figure*}[t]
    \centering
    \includegraphics[width=0.98\textwidth]{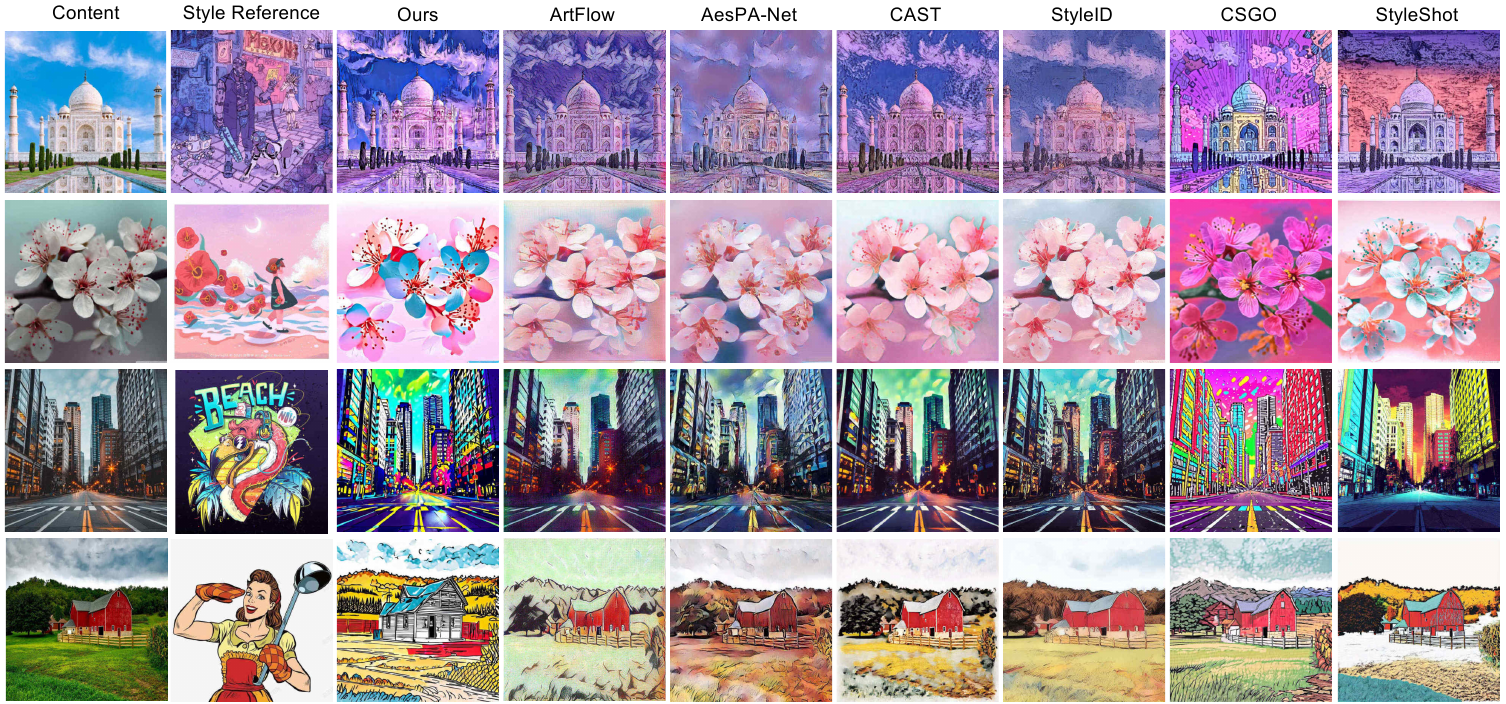}
    \caption{Qualitative comparison with other state-of-the-art methods for the image-guided style transfer task. }
    
    \label{fig:img_compare}
\end{figure*}

\subsection{Qualitative Experiments}

\noindent{\textbf{Instruction-Guided Style Transfer}}. As shown in Figure \ref{fig:instruction_compare}, we conducted a comprehensive comparison with five SOTA methods across multiple style categories. Current text-guided style transfer methods, such as DiffStyler, are limited in their ability to handle diverse style categories. For instance, DiffStyler struggles with fine-grained styles, such as American comic, and may fail to preserve the original content, as seen with images like Elon Musk. On the other hand, instruction-guided image editing methods, represented by OmniGen and UltraEdit, are also constrained in style transfer tasks. These methods cannot fully understand fine-grained style information, which can lead to significant content alterations (e.g., HQ-Edit) or insufficient stylization (e.g., OmniGen, UltraEdit). In contrast, OmniStyle benefits from training on a large-scale, high-quality dataset of stylized triplets, allowing it to learn rich, fine-grained style information. As a result, OmniStyle is better equipped to handle a broader range of style categories and produce more accurate and consistent results.

\noindent{\textbf{Image-Guided Style Transfer}}. 
As shown in Figure \ref{fig:img_compare}, we performed a comprehensive comparison with six SOTA image-guided style transfer methods. In comparison, OmniStyle demonstrates superior style preservation (see row 3), enhanced aesthetic appeal (see row 4), and better content preservation (see row 1).  More results are provided in supplementary materials.

% \subsection{StyleFilter VS Traditional Evaluation Methods }

% 定量
% 定性
% User Study

% \vspace{-5pt}
\subsection{User Study}
We conducted a user study to evaluate the effectiveness of our method in instruction-guided and image-guided style transfer tasks. Participants were asked to rank their top three preferred results based on the following criteria: (1) \textbf{Style Preservation}, the degree to which the generated image retains the stylistic characteristics of the given style image; (2) \textbf{Content Preservation}, how well the structural details of the content image are maintained; (3) \textbf{Aesthetic Appeal}, the overall visual quality of the generated image. To reduce bias, options were presented in a randomized order, and participants were allowed to zoom in for closer inspection. A total of 1,800 votes were collected from 30 participants. As demonstrated in Table~\ref{tab:user_1} and Table~\ref{tab:user_2}, we computed the proportion of first-place rankings (Rank 1). These results highlight that the stylized images generated by our method were preferred by users, showcasing its superior performance in style transfer. More detailed results and analyses are provided in the supplementary material.

\begin{table}[]
    \caption{The user study for instruction-guided style transfer tasks.}
    \label{tab:user_1}
    \resizebox{\linewidth}{!}{%
    \setlength{\tabcolsep}{2.5pt}
    \begin{tabular}{ccccccc}
    \hline
    Methods/Metrics & OmniStyle & DiffStyler & HQ-Edit & UltraEdit & Instruct-Pix2Pix & OmniGen \\ \hline
    Rank 1          &  \textbf{86.90\%}   &         1.19\%   &     0.40\%    &         5.16\%  &    1.19\%              &      5.16\%    \\ \hline
    \end{tabular}
    }
    \vspace{-1.5em} 
    \end{table}

    \begin{table}[]
    \caption{The user study for image-guided style transfer tasks.}
    \label{tab:user_2}
    \resizebox{\linewidth}{!}{%
    \setlength{\tabcolsep}{2.5pt}
    \begin{tabular}{cccccccc}
    \hline
    Methods/Metrics & OmniStyle & ArtFlow & AesPA-Net & CAST & StyleID & CSGO & StyleShot \\ \hline
    Rank 1          &    \textbf{41.22\% }      &    2.87\%     &     3.94\%      &   10.75\%   &    7.89\%     &   14.70\%   &     18.63\%      \\ \hline
    \end{tabular}
    }
    \vspace{-1.5em} 
    \end{table}

\section{Conclusion}
\label{conclusion}
In this paper, we addressed key limitations in style transfer by introducing OmniStyle-1M, a large-scale, high-quality dataset of one million content-style-stylized image triplets across 1,000 style categories. This paired dataset, enriched with detailed prompts and descriptions, enables models to achieve more controlled, consistent, and diverse style transfer outcomes, moving beyond the limitations of arbitrary style transfer. To ensure the quality of stylized outputs, we proposed OmniFilter, which is designed to assess and filter images based on content preservation, style consistency, and aesthetic appeal. This filtering process ensures that OmniStyle-1M serves as a robust basis for training high-quality style transfer models. Last, we introduced OmniStyle, an end-to-end style transfer framework based on the DiT architecture. OmniStyle supports both instruction-guided and image-guided tasks, achieving high-resolution and detailed stylizations. Through extensive experiments, we demonstrated that OmniStyle outperforms existing methods in both quality and efficiency, establishing it as a strong baseline for future research and real-world applications. Together, OmniStyle-1M, OmniFilter, and OmniStyle lay a comprehensive foundation for advancing style transfer, contributing valuable resources and methodologies for the research community. 

In the future, a potential direction is to  collaborate with professional artist to annotate perfect paired dataset within OmniStyle-1M dataset. Such perfect paired datasets provide a reference ``ground truth" stylized image for each content-style pair, which allows for more objective evaluation of a model’s performance. With paired data, metrics like perceptual similarity or content preservation can be accurately measured by comparing generated images to the target, helping to assess style accuracy and consistency more rigorously.

\section*{Acknowledgments}
This work  was supported in part by the National Natural Science Foundation of China (No.\ 62202199, 62406134), the Suzhou Key Technologies Project (No. SYG2024136) and the Fundamental Research Funds for the Central Universities.

\normalem
{
    \small
    \bibliographystyle{ieeenat_fullname}
    \bibliography{main}
}

\newpage

% WARNING: do not forget to delete the supplementary pages from your submission 
% CVPR 2025 Paper Template; see https://github.com/cvpr-org/author-kit
\setcounter{section}{0}
\twocolumn[{%
\begin{center}
    {\LARGE \textbf{Supplementary Materials}}\\[0.5em]
\end{center}

\begin{center}
    \includegraphics[width=1.0\textwidth]{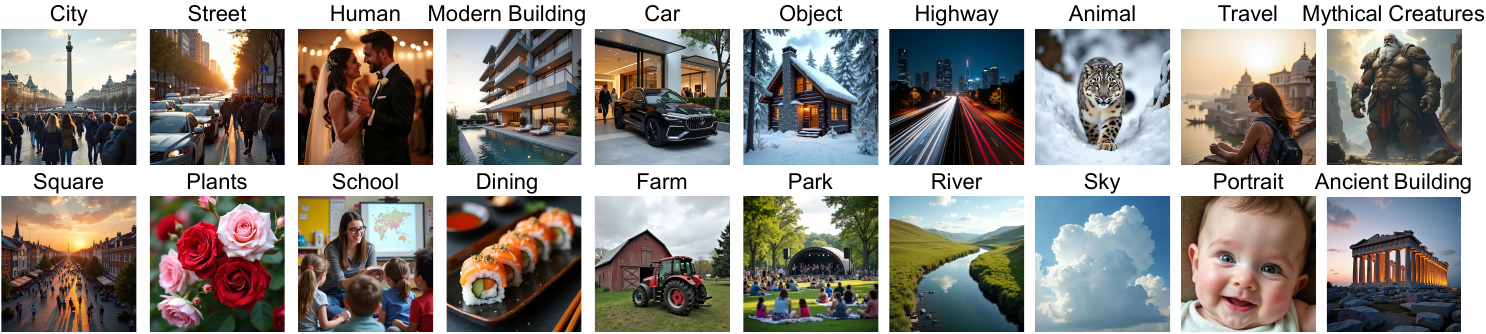}
    \captionof{figure}{Examples of content images across 20 categories. Each image is randomly selected from its respective category, showcasing the diversity of the categories and the high quality of the generated images.}
    \label{fig:cnt_20}
\end{center}
}]

\renewcommand{\thefootnote}{\fnsymbol{footnote}}
\footnotetext[1]{Corresponding authors.}
\renewcommand{\thefootnote}{\arabic{footnote}} % 恢复正常编号格式
% Please add the following required packages to your document preamble:

% \begin{figure*}[t]
%     \centering
%     \includegraphics[width=1.0\textwidth]{assets/cnt_overview.pdf}
%     \caption{Examples of content images across 20 categories. Each image is randomly selected from its respective category, showcasing the diversity of the categories and the high quality of the generated images.}
%     \label{fig:cnt_20}
% \end{figure*}

% \begin{table*}[]
% \caption{The information of the 20 categories of content images.}
% \label{tab:cnt_class}
% \resizebox{\textwidth}{!}{%
% \begin{tabular}{cccccccccc}
% \hline
% City   & Street & Human  & Modern Building & Car  & Object & Highway & Animal & Travel   & Mythical Creatures   \\
% Square & Plant  & School & Dining          & Farm & Park   & River   & Sky    & Portrait & Ancient Building \\ \hline
% \end{tabular}%
% }
% \end{table*}

\section{More Details about OmniStyle}
In this section, we provide more details about OmniStyle. Section \ref{cnt_sec} delves into the process of generating content images, while Section \ref{sty_sec} elaborates on the specific details of the collected style images. More triplet examples from our dataset are presented in Section \ref{sec_tri}.

\subsection{Content Images Generation}
\label{cnt_sec}
We adopt a generative approach to obtain content images, which offers the advantages of effectively avoiding issues related to copyright and privacy while ensuring high-quality data. 
Specifically, we first define 20 common image categories
%, as shown in Table \ref{tab:cnt_class}. 
For each category, we utilize ChatGPT \cite{openai2024chatgpt} to automatically generate 100 textual descriptions. Based on these descriptions, we employ FLUX \cite{flux} to generate the corresponding images, resulting in a total of 2,000 content images across the 20 categories.
Figure \ref{fig:cnt_20} provides an overview of all 20 categories and the exemplar generated content images.
Each image is rendered at a 1K resolution, guaranteeing high visual clarity and fidelity. 
Table \ref{tab:text_prompt} presents the some exemplar textual prompts for generating the animal category as a reference.

% Please add the following required packages to your document preamble:
% \usepackage{graphicx}
% \usepackage[normalem]{ulem}
% \useunder{\uline}{\ul}{}

% \noindent{\textbf{Prompts}}
% img展示

\begin{table}[t]
\caption{Ten textual prompts for the animal category.}
\label{tab:text_prompt}
\resizebox{\columnwidth}{!}{%
\begin{tabular}{lllllllllllllll}
\hline
\multicolumn{15}{l}{1. A swan swimming gracefully on a calm lake at sunset.}     \\
\multicolumn{15}{l}{2. A leopard resting on the branch of a tree in the jungle.} \\
\multicolumn{15}{l}{3. A close-up of a ladybug crawling on a green leaf.}        \\
\multicolumn{15}{l}{4. A family of deer grazing in a peaceful meadow.}           \\
\multicolumn{15}{l}{5. A family of ducks swimming in a calm pond.}               \\
\multicolumn{15}{l}{6. A polar bear swimming in the icy waters of the Arctic.}   \\
\multicolumn{15}{l}{7. A family of ducks swimming in a calm pond.}               \\
\multicolumn{15}{l}{8. A majestic lion standing on a rocky ledge at sunset.}     \\
\multicolumn{15}{l}{9. A hawk swooping down to catch its prey.}                  \\
\multicolumn{15}{l}{10. A close-up of a spider spinning a web.}                  \\ \hline
\end{tabular}%
}
\end{table}

\begin{figure*}[t]
    \centering
    \includegraphics[width=1.0\textwidth]{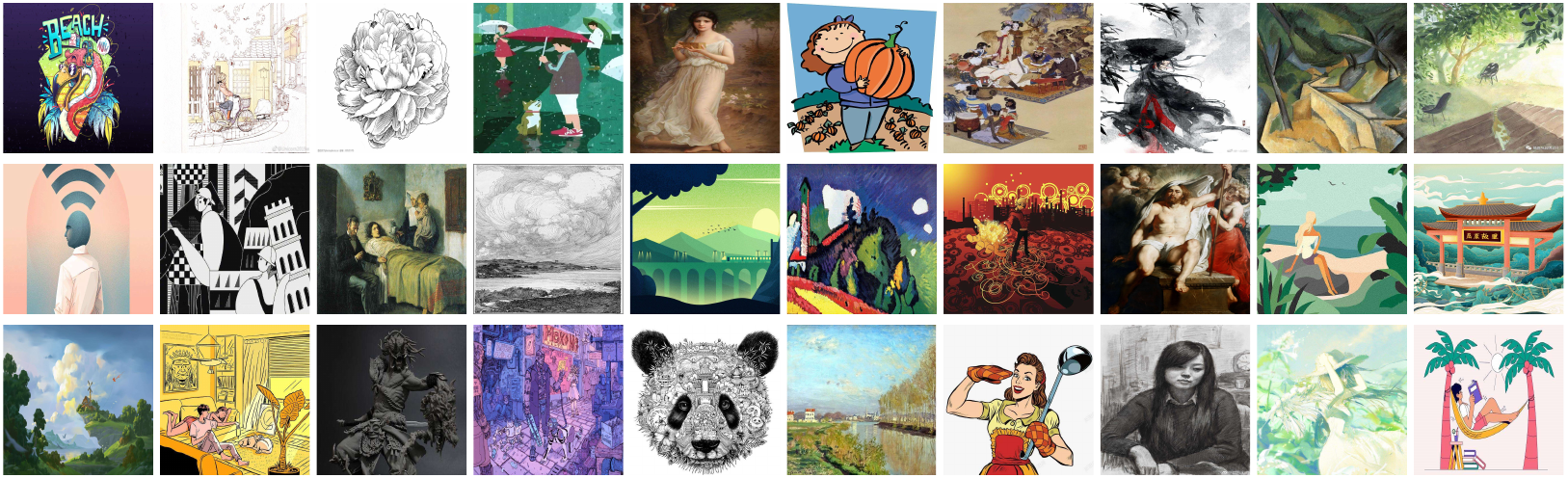}
    \caption{A subset of style images in OmniStyle Dataset.}
    \label{fig:sty_overview}
\end{figure*}

\begin{figure*}[t]
    \centering
    \includegraphics[width=1.0\textwidth]{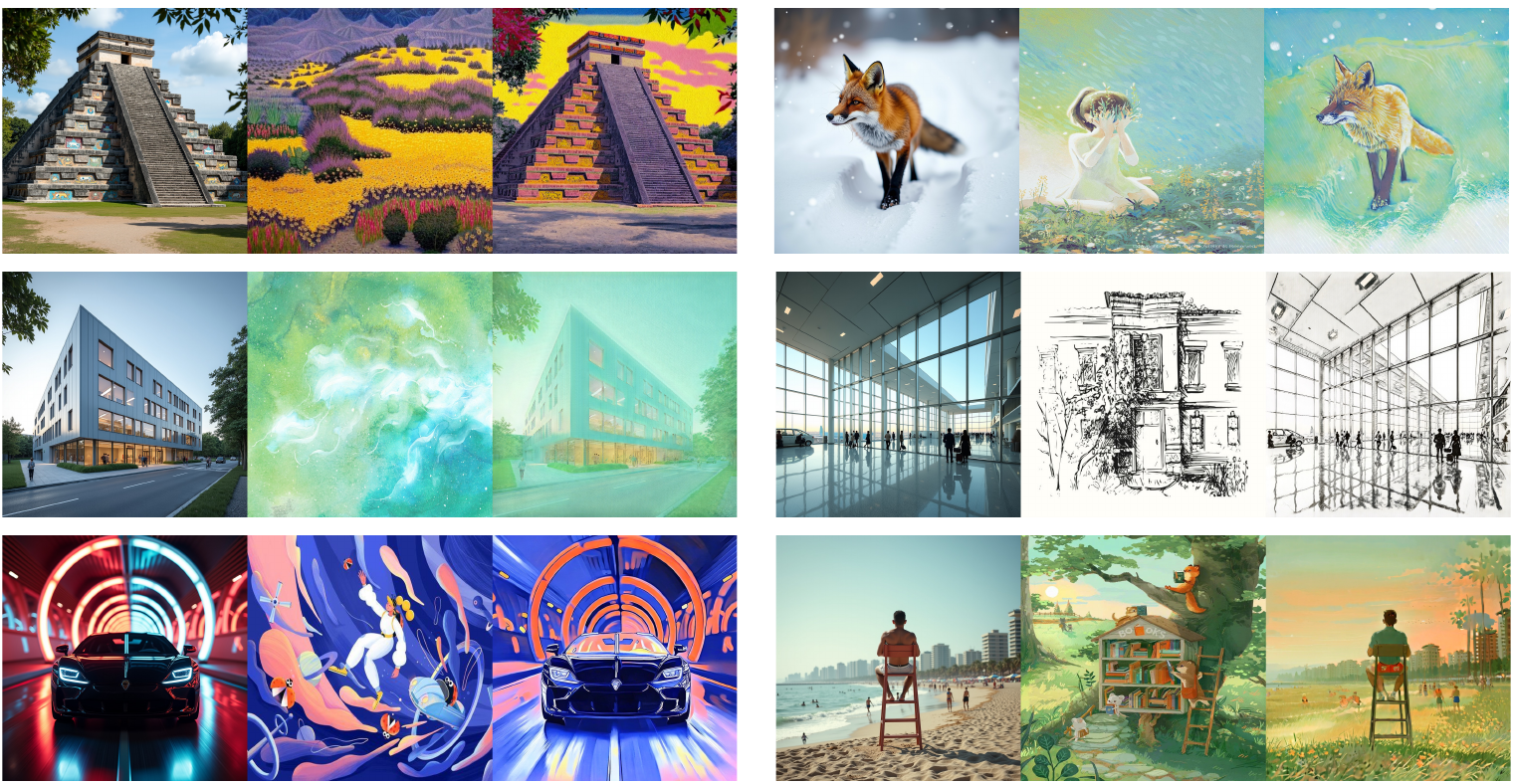}
    \caption{Examples of Triplet from the OmniStyle-1M dataset. From left to right, each set of images is arranged as: content image, style image, and resulting stylized image.}
    \label{fig:triplets}
\end{figure*}

\begin{figure*}[t]
    \centering
    \includegraphics[width=1.0\textwidth]{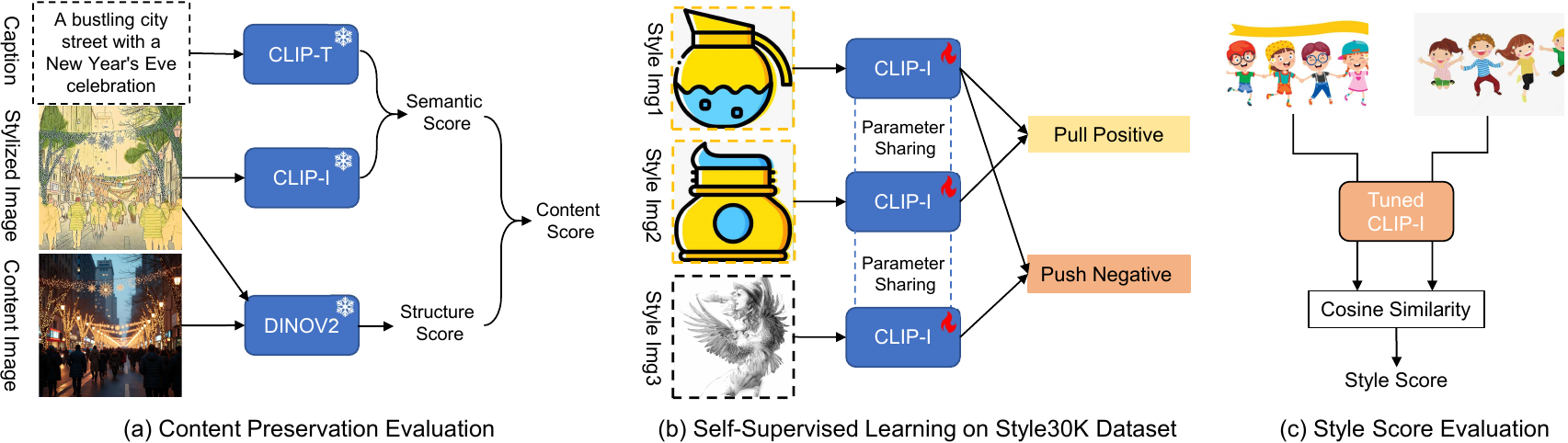}
    \caption{Frameworks for content preservation evaluation and style consistency evaluation.}
    \label{fig:cnt_sty_score}
\end{figure*}

\begin{figure*}[t]
    \centering
    \includegraphics[width=1.0\textwidth]{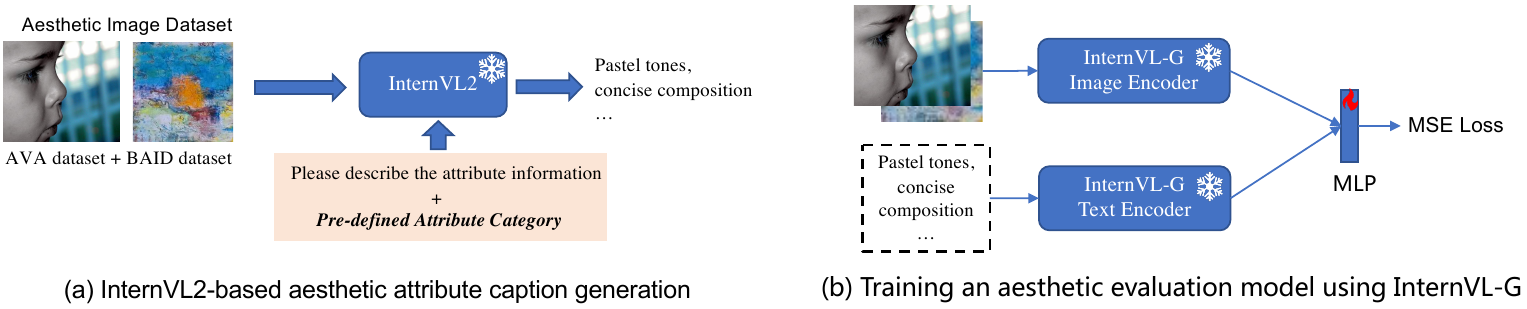}
    \caption{Frameworks for aesthetic appeal evaluation.}
    \label{fig:aes_score}
\end{figure*}

\subsection{Style Images Collections}
\label{sty_sec}
Existing style image datasets typically focus on a limited range of traditional style categories, such as oil painting, watercolor, and sketches. In contrast, our study highlights a broader and more fine-grained spectrum of style categories, encompassing up to 1,000 distinct styles collected from the Style30K \cite{li2024styletokenizer} dataset. In Figure \ref{fig:sty_overview}, we provide examples of some style images for reference.

\subsection{Examples of Triplet }
\label{sec_tri}
We provide additional examples of triplet from the OmniStyle-1M dataset, as illustrated in Figure \ref{fig:triplets}. Both the content images and the generated stylized images used in this study maintain a resolution of 1024*1024, ensuring high-quality data. Moreover, it can be found the stylized result images exhibit a high degree of style consistency with the designated reference styles, which reflects the superiority of our constructed dataset in preserving style.

\section{More Details about OmniFilter}

In this section, we provide more details about the evaluation schemes for content preservation, style consistency and aesthetic appeal, which are the three key aspects of the OmniFilter.

\noindent{\textbf{Evaluation on Content Preservation.}}
We present the detailed model architecture for content preservation evaluation  as shown in Figure \ref{fig:cnt_sty_score}.a. In this architecture, CLIP \cite{radford2021learning} is employed to compute the similarity score between the stylized image and the content text, serving as a measure of semantic similarity. Simultaneously, DINOV2 \cite{oquab2023dinov2} is used to calculate the structural similarity between the stylized image and the original content image. This process does not require any training or fine-tuning. Specifically, the pre-trained weights for CLIP \cite{radford2021learning} are from ``clip-vit-large-patch14," and for DINOV2 \cite{oquab2023dinov2}, ``dinov2-large" is utilized.

\noindent{\textbf{Evaluation on Style Consistency.}}
As illustrated in Figures \ref{fig:cnt_sty_score}.b and c, we present the training and inference architecture for style consistency evaluation. In this architecture, we adopt a contrastive learning approach, leveraging the Style30K \cite{li2024styletokenizer} dataset to construct positive and negative sample pairs for updating the CLIP \cite{radford2021learning} image encoder. Once training is completed, the fine-tuned CLIP \cite{radford2021learning} image encoder can be used to assess the similarity between any two style images.

\begin{table}[t]
\caption{Comparison between our method and the original CLIP model on the style image retrieval task.}
\label{tab:sty_compare_supp}
%\resizebox{\columnwidth}{!}{%
\centering
\begin{tabular}{cccc}
\hline
Models & Rank1            & Rank5            & Rank10           \\ \hline
CLIP \cite{radford2021learning}   & 15.00\%          & 34.23\%          & 42.50\%          \\
Ours   & \textbf{40.19\%} & \textbf{65.96\%} & \textbf{78.46\%} \\ \hline
\end{tabular}%
%}
\end{table}

 To further demonstrate the effectiveness of our method, we randomly selected 500 images from the Style30K \cite{li2024styletokenizer} dataset as queries and another 500 corresponding images as keys. These images did not appear in the training data.
As shown in Table \ref{tab:sty_compare_supp}, our model, which is fine-tuned by self-supervised learning, outperforms the original CLIP model in style retrieval tasks. This demonstrates that our method is better suited for calculating style similarity, further confirming its effectiveness.

\begin{figure}[t]
    \centering
    \includegraphics[width=0.9\columnwidth]{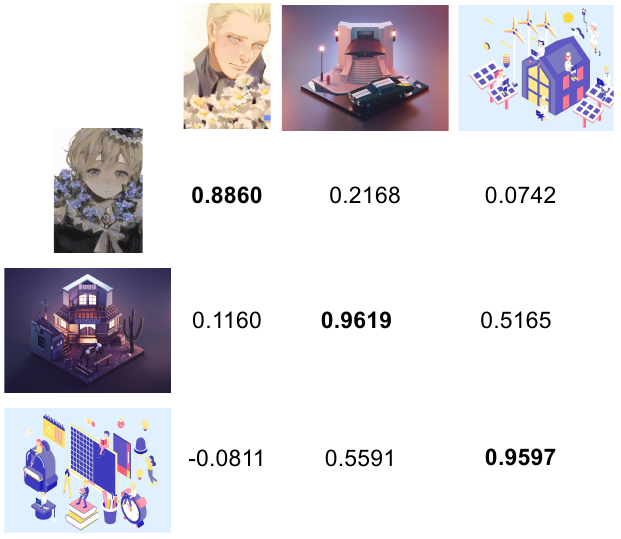}
    \caption{The similarity evaluation results for different style image pairs using our model, with high scores highlighted in bold indicating greater style consistency.}
    \label{fig:sty_results}
\end{figure}

Figure \ref{fig:sty_results} illustrates the similarity evaluation results of different style image pairs using our model. Each row represents a group of style image pairs, with the computed similarity scores displayed accordingly. High scores (highlighted in bold) indicate greater style consistency between the image pairs, demonstrating the effectiveness of our method in capturing style similarity.

\begin{figure*}[t]
    \centering
    \includegraphics[width=1.0\textwidth]{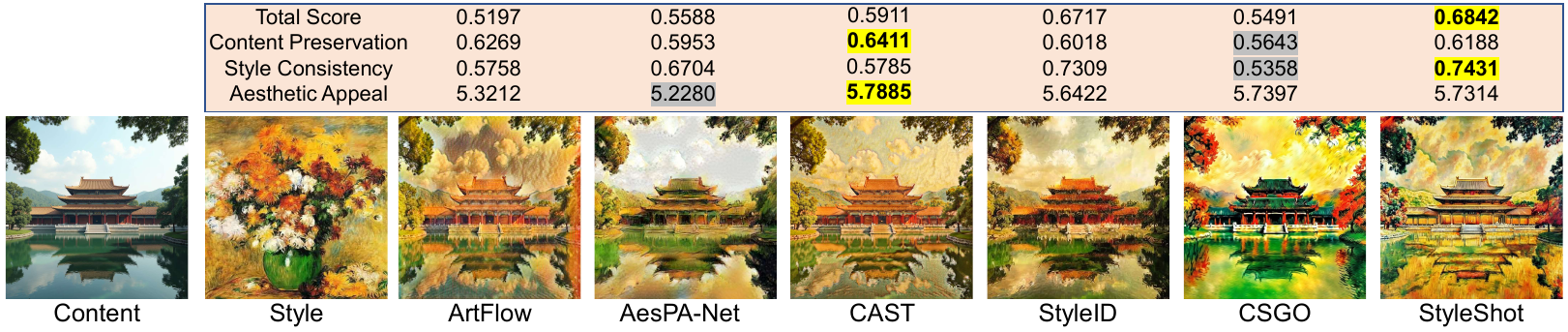}
    \caption{Qualitative ablation study results for the core components of OmniFilter. The yellow highlights indicate the highest values, while the gray highlights indicate the lowest values.}
    \label{fig:ablation}
\end{figure*}

\noindent{\textbf{Evaluation Aesthetic Appeal.}}
The detailed framework for aesthetic appeal evaluation is shown in Figure \ref{fig:aes_score}. In subfigure (a), we utilize InternVL2 \cite{chen2023internvl} to generate attribute captions for existing aesthetic image datasets, including the natural image dataset AVA \cite{murray2012ava} and the artistic style image dataset BAID \cite{Yi_2023_CVPR}. Then, the predefined text templates covering 40 visual attribute categories are used as query, requiring the model to describe various attributes of the current image. After obtaining the visual attribute descriptions, InternVL-G \cite{chen2023internvl} is employed to extract features from both text and images as shown in Figure \ref{fig:aes_score}.b. Subsequently, an MLP layer is trained to predict aesthetic scores. Here, the 1B version of InternVL2 is used, while the pre-trained weights of InternVL-G are based on InternVL-14B-224px. The predefined text prompt is shown in Table \ref{tab:text_att}.

\begin{table*}[]
\caption{The predefined text prompt for visual attributes caption generation.}
\label{tab:text_att}
\resizebox{\textwidth}{!}{%
\begin{tabular}{c}
\hline
Please provide a comprehensive description of the image based on  the following visual attributes: Focus on how each attribute contributes to \\ 
the overall aesthetic quality of the image, including aspects such as composition, balance, color harmony, lighting, contrast, texture, \\ 
sharpness, depth of field, simplicity, symmetry and patterns, framing, leading lines, rule of thirds, perspective, emotional impact, uniqueness, \\ 
visual storytelling, rhythm, harmony, proportion, gradation, tonal range, vibrance, clarity, negative space, detail, mood, geometric elements, \\ 
visual hierarchy, light quality, visual tension, focus and depth, cultural significance, form and shape, saturation, motion blur, \\ 
point of view (POV), symbolism, consistency, and foreground and background interaction. Summarize it in one paragraph.  \\ \hline
\end{tabular}%
}
\end{table*}

\section{More Experiments}
In this section, we present additional experimental results. Section \ref{quan_exp} focuses on quantitative evaluations, including user studies and ablation studies. Section \ref{qual_exp} highlights more qualitative results for further analysis.

\subsection{Quantitative Experiments}
\label{quan_exp}

\noindent{\textbf{User Study.}}
For the instruction-guided and image-guided style transfer tasks, participants were asked to select the top 3 results and rank them based on the following evaluation criteria:  
\begin{enumerate}[label=(\arabic*)]
    \item \textbf{Style Preservation}: How well does the generated result retain the stylistic characteristics of the given style image?  
    \item \textbf{Content Preservation}: How effectively does the generated result preserve the content and structural details of the given content image?  
    \item \textbf{Aesthetic Appeal}: How is the overall visual quality and aesthetic appeal of the generated result?  
\end{enumerate}  
The options were presented in a randomized order to mitigate potential bias. To ensure usability, participants could zoom in the images for closer examination.  

As shown in Tables \ref{tab:user_1_supp} and \ref{tab:user_2_supp}, we provide detailed results of the user study, including Rank 1 and Top 3 scores. Our method, OmniStyle, achieves the highest user ratings in both instruction-guided and image-guided style transfer tasks, demonstrating its effectiveness in generating user-preferred results.

\begin{table}[t]
\caption{The user study for instruction-guided style transfer tasks.}
\label{tab:user_1_supp}
\resizebox{\linewidth}{!}{%
\setlength{\tabcolsep}{1.6pt}
\begin{tabular}{ccccccc}
\hline
Methods/Metrics & OmniStyle & DiffStyler & HQ-Edit & UltraEdit & Instruct-Pix2Pix & OmniGen \\ \hline
Rank 1          &  \textbf{86.90\%}   &         1.19\%   &     0.40\%    &         5.16\%  &    1.19\%              &      5.16\%    \\ 
Top 3          &  \textbf{98.4\%}   &         34.23\%   &     5.13\%    &         59.1\%  &    49.59\%              &      53.55\%    \\ \hline

\end{tabular}
}
\end{table}

\begin{table}[t]
\caption{The user study for image-guided style transfer tasks.}
\label{tab:user_2_supp}
\resizebox{\linewidth}{!}{%
\setlength{\tabcolsep}{1.6pt}
\begin{tabular}{cccccccc}
\hline
Methods/Metrics & OmniStyle & ArtFlow & AesPA-Net & CAST & StyleID & CSGO & StyleShot \\ \hline
Rank 1          &    \textbf{41.22\% }      &    2.87\%     &     3.94\%      &   10.75\%   &    7.89\%     &   14.70\%   &     18.63\%      \\ 
Top 3          &    \textbf{59.34\% }      &    22.65\%     &     38.46\%      &   46.02\%   &    41.85\%     &  45.30\%   &     46.38\%      \\ \hline
\end{tabular}
}
\end{table}

\begin{table}[t]
\caption{Ablation Study for the core components of OmniFilter.}
\label{tab:ablation_supp}
\resizebox{\columnwidth}{!}{%
\begin{tabular}{cc}
\hline
Components                                              & Style Loss $\downarrow$ \\ \hline
Content Preservation                                    & 0.1577           \\
Content Preservation+Style Consistency                  & 0.1434           \\
Content Preservation+Style Consistency+Aesthetic Appeal & \textbf{0.1405}           \\ \hline
\end{tabular}%
}
\end{table}

\begin{figure*}[t]
    \centering
    \includegraphics[width=1.0\textwidth]{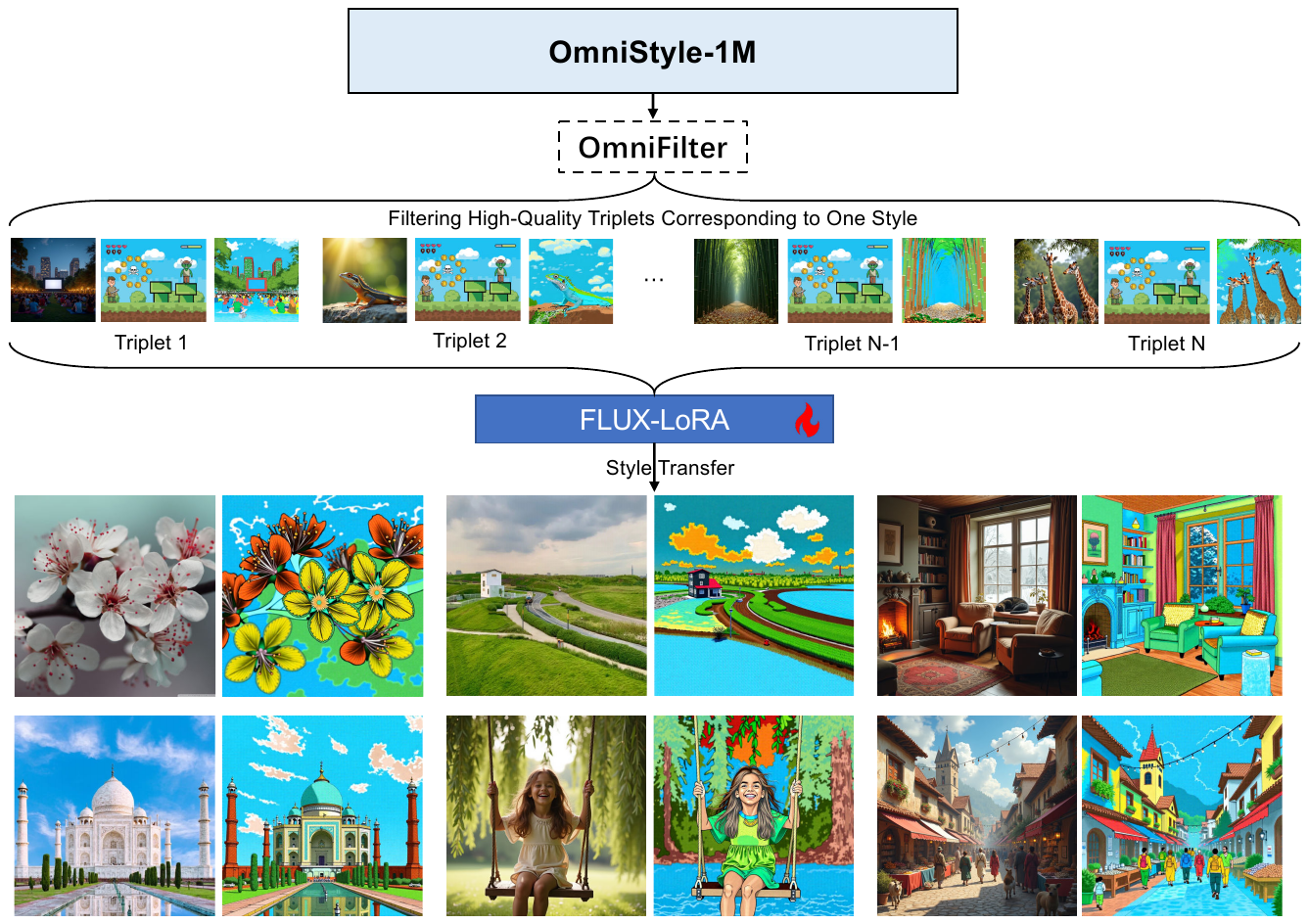}
    \caption{The data filtering and model training pipeline for LoRA-based style transfer task.}
    \label{fig:lora}
\end{figure*}

\noindent{\textbf{Ablation Study.}}
Table \ref{tab:ablation_supp} presents the results of the ablation study evaluating the contributions of the core components of the OmniFilter framework: Content Preservation, Style Consistency, and Aesthetic Appeal. We filtered the OmniStyle-1M according to the strategies outlined in the table and evaluated the filtering results using the Style Loss metric, where lower values indicate better data quality.

As shown in the table, the inclusion of Style Consistency alongside Content Preservation further improves performance, reducing the Style Loss from 0.1577 to 0.1434. This highlights the pivotal role of Style Consistency in aligning the generated image with the target style. Furthermore, incorporating the Aesthetic Appeal component on top of the previous two achieves the best performance, with a Style Loss of 0.1405. This improvement underscores the importance of considering aesthetic factors for generating high-quality style transfer results.

Furthermore, as shown in Figure \ref{fig:ablation}, we present a qualitative analysis of each component of OmniFilter. In terms of content preservation, CAST \cite{zhang2020cast} achieves the highest score (0.6411) due to its superior ability to retain the structure and texture of the content image (e.g., the clouds). For style similarity, StyleShot \cite{gao2024styleshot} attains the highest score (0.7431), while CSGO \cite{xing2024csgo} produces results with significantly different color distributions, leading to the lowest score (0.5358). From an aesthetic perspective, ArtFlow \cite{artflow2021} and AesPA-Net \cite{hong2023aespa} exhibit lower aesthetic scores due to the presence of blurred or shadowed regions in their results (visible upon magnification). In contrast, the outputs generated by CAST, CSGO, and StyleShot display rich details, resulting in relatively higher aesthetic scores. By comprehensively considering all three aspects, StyleShot achieves the highest overall score. These findings demonstrate that OmniFilter effectively balances and evaluates style transfer outcomes, enabling high-quality data filtering.

In summary, the progressive integration of these components illustrates their complementary effects, thereby enhancing the overall performance of the OmniFilter framework.

% 定性
\subsection{Qualitative Experiments}
\label{qual_exp}

\noindent{\textbf{Lora-Based Style Transfer}}. 
Thanks to the inclusion of hundreds of triplets for each style category in the OmniStyle-1M dataset, we are able to use OmniFilter to extract a high-quality subset of triplets corresponding to the target style. This subset can effectively support the training of LoRA models. As shown in Figure \ref{fig:lora}, we illustrate the LoRA training pipeline based on OmniStyle-1M and OmniFilter. In this process, we adopt the official FLUX-LoRA training method, with the sole difference being that, to achieve style transfer, we encode the content image using a VAE and concatenate it with noise latents before feeding them into the model. Experimental results demonstrate that the LoRA model trained on our dataset can achieve high-quality style transfer, thus validating the significance of our dataset and the effectiveness of our filtering method.

% \item  \textbf{Image-Based Style Transfer}. In Figures \ref{fig:res12}–\ref{fig:res910}, we provide more results of image-guided style transfer. 
%     \item  \textbf{Instruction-Based Style Transfer}. In Figures \ref{fig:res1112}, we provide more results of instruction-guided style transfer. 

\begin{figure*}
    \centering
    \includegraphics[width=1\textwidth]{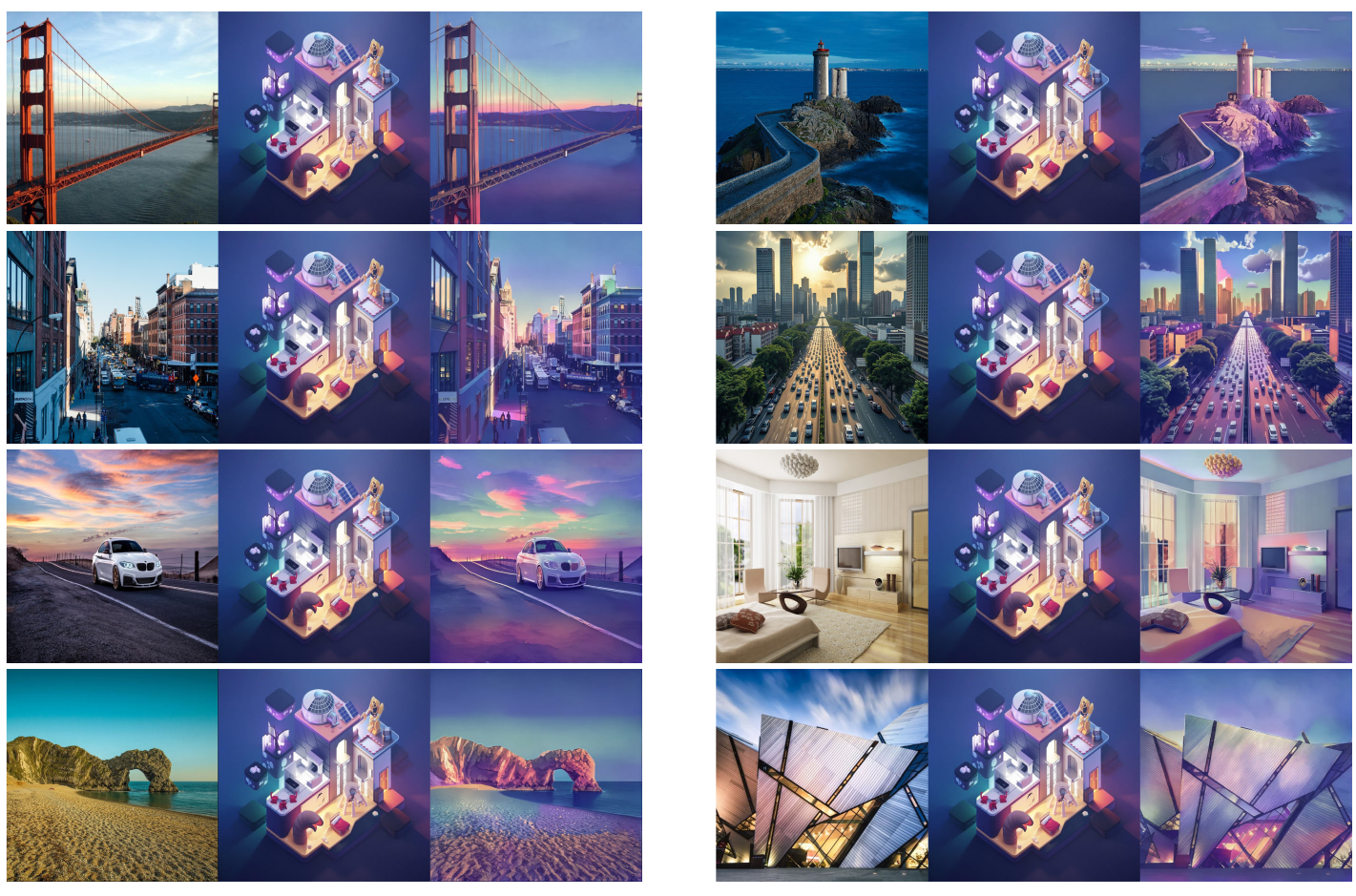}
    \includegraphics[width=1\textwidth]{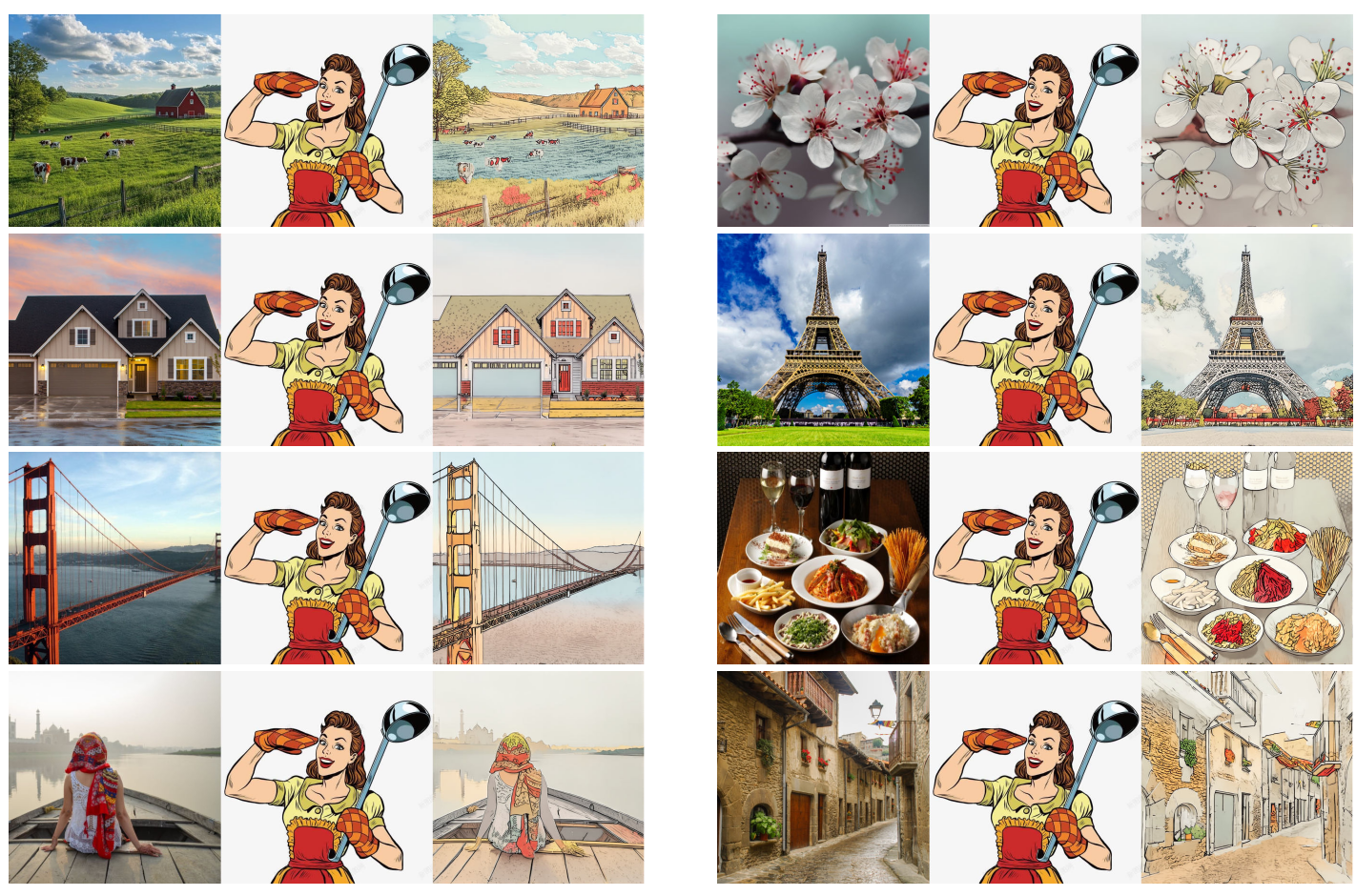}

    \caption{Additional image-guided style transfer results of OmniStyle. Left: content image, middle: style reference, right: stylized output.}
    \label{fig:res12}
\end{figure*}

\begin{figure*}
    \centering
    \includegraphics[width=1\textwidth]{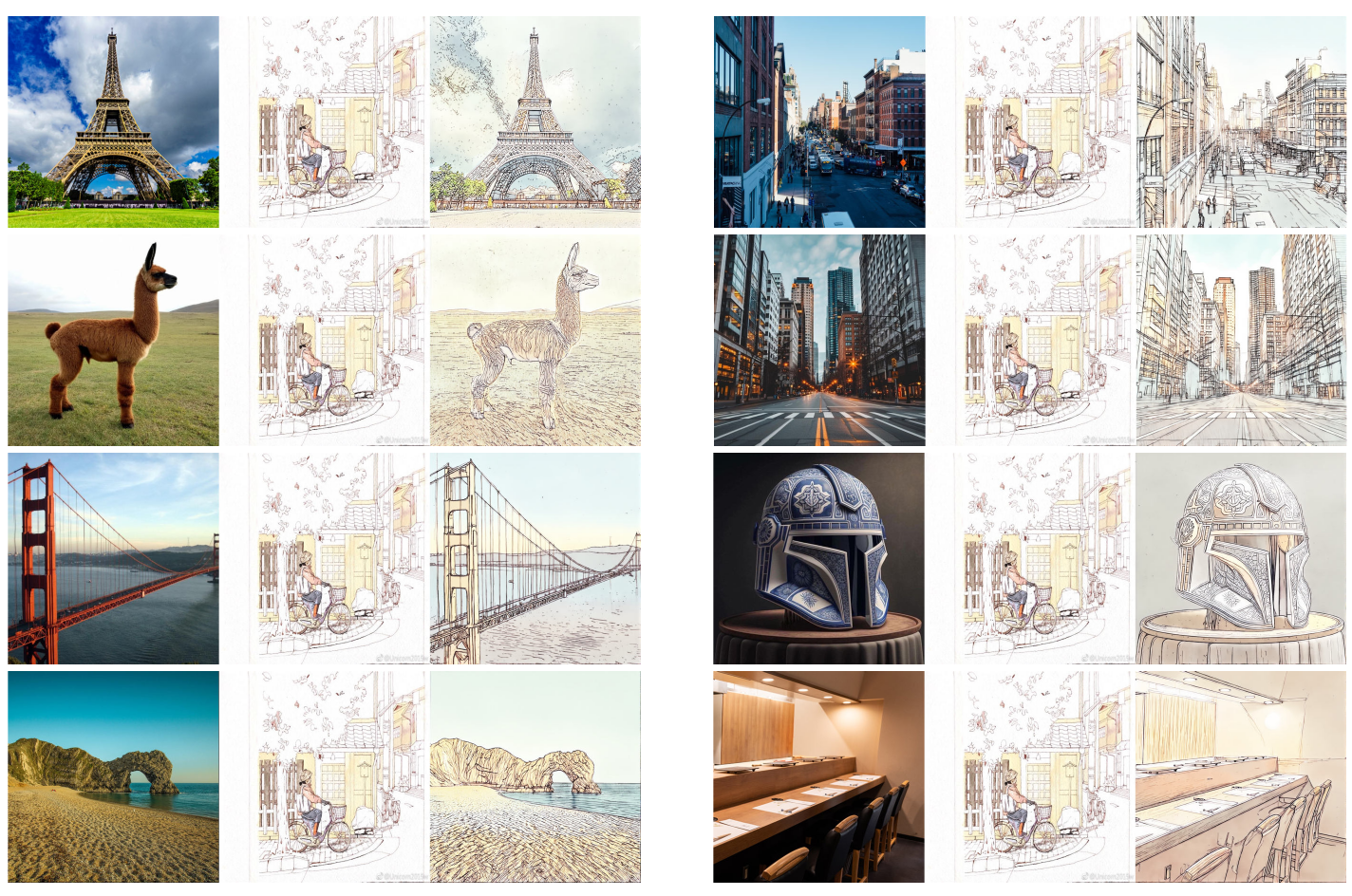}
    \includegraphics[width=1\textwidth]{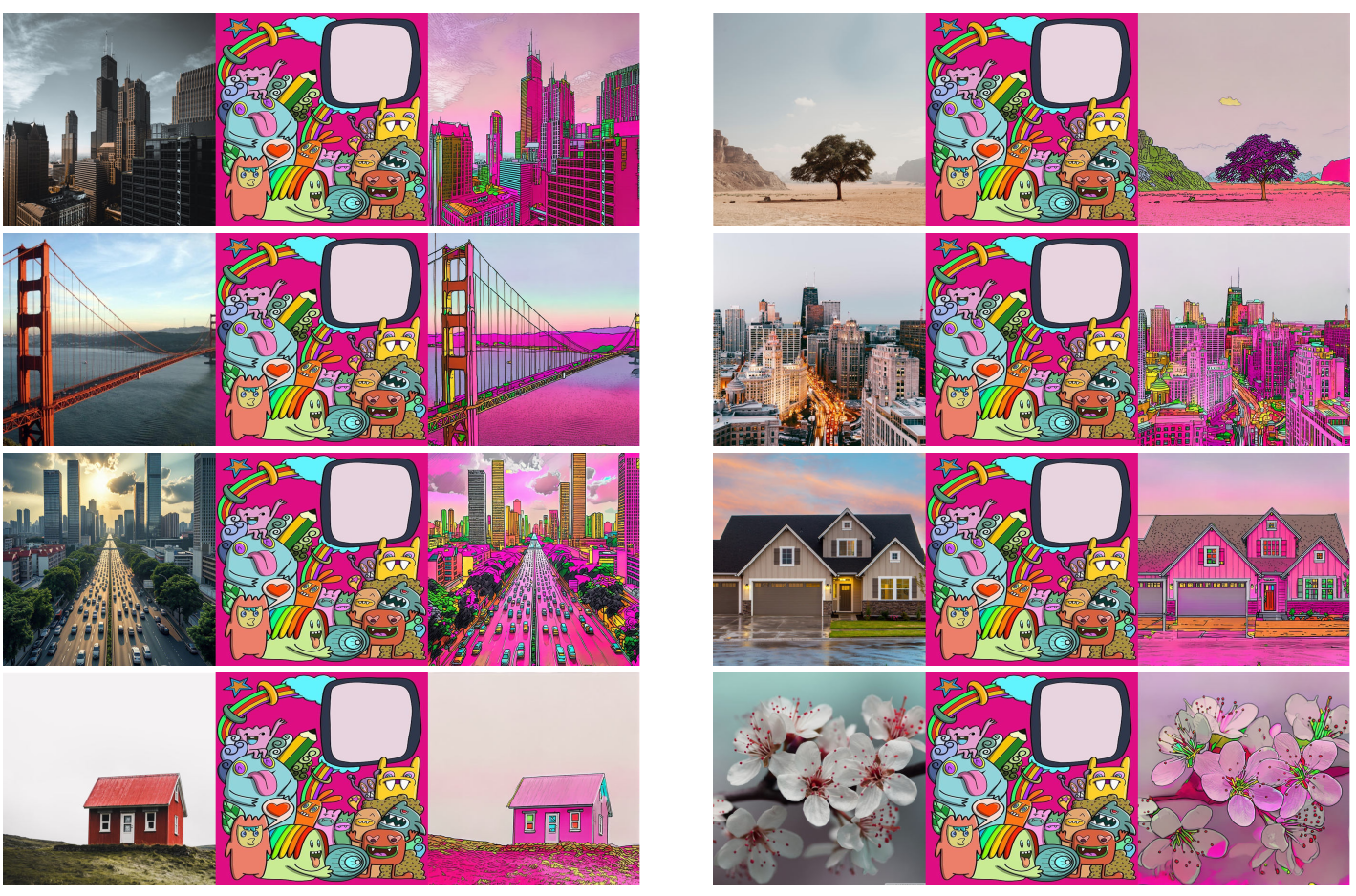}

    \caption{Additional image-guided style transfer results of OmniStyle. Left: content image, middle: style reference, right: stylized output.}
    \label{fig:res34}
\end{figure*}

\begin{figure*}
    \centering
    \includegraphics[width=1\textwidth]{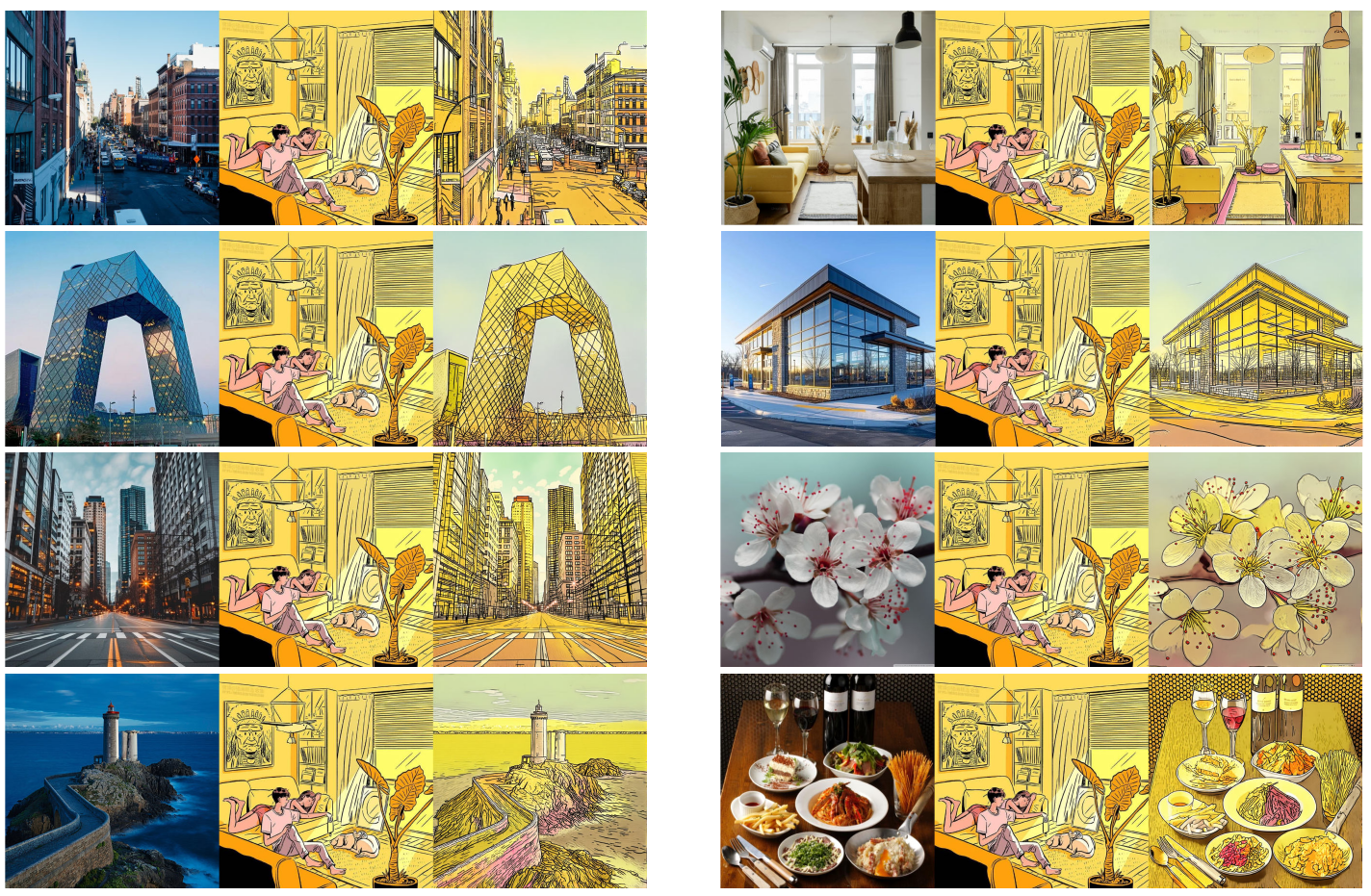}
    \includegraphics[width=1\textwidth]{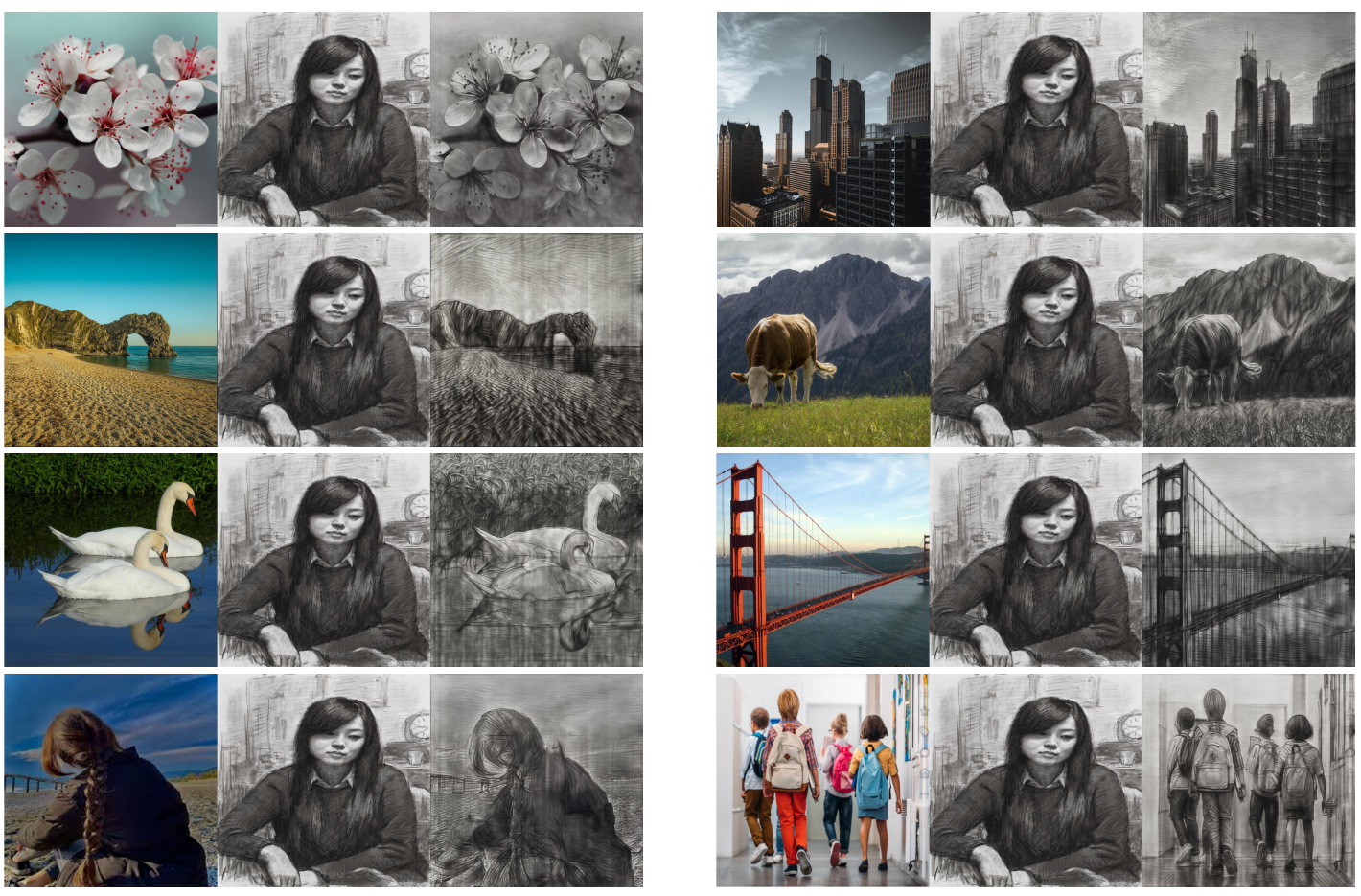}

    \caption{Additional image-guided style transfer results of OmniStyle. Left: content image, middle: style reference, right: stylized output.}
    \label{fig:res56}
\end{figure*}

\begin{figure*}
    \centering
    \includegraphics[width=1\textwidth]{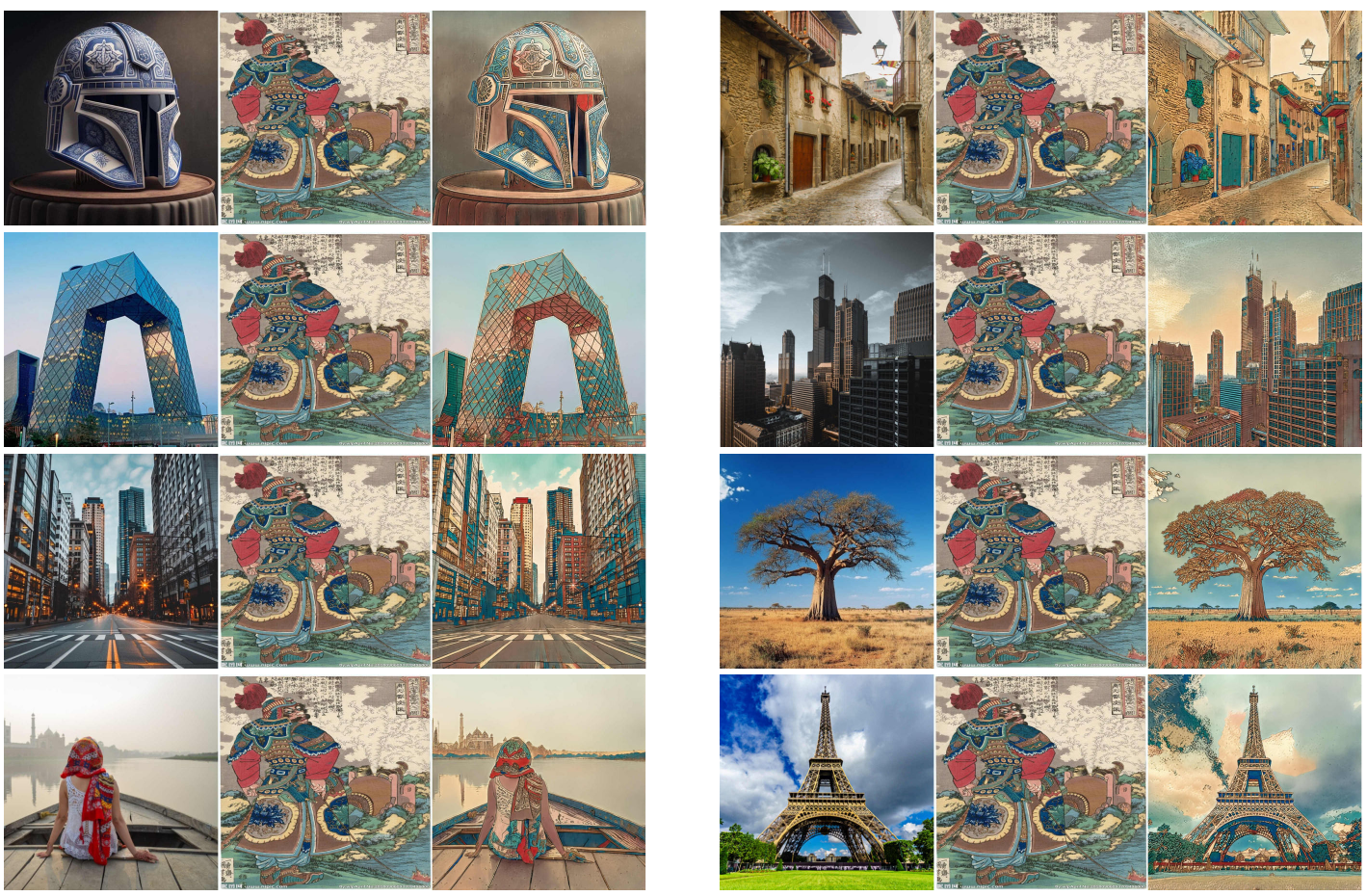}
    \includegraphics[width=1\textwidth]{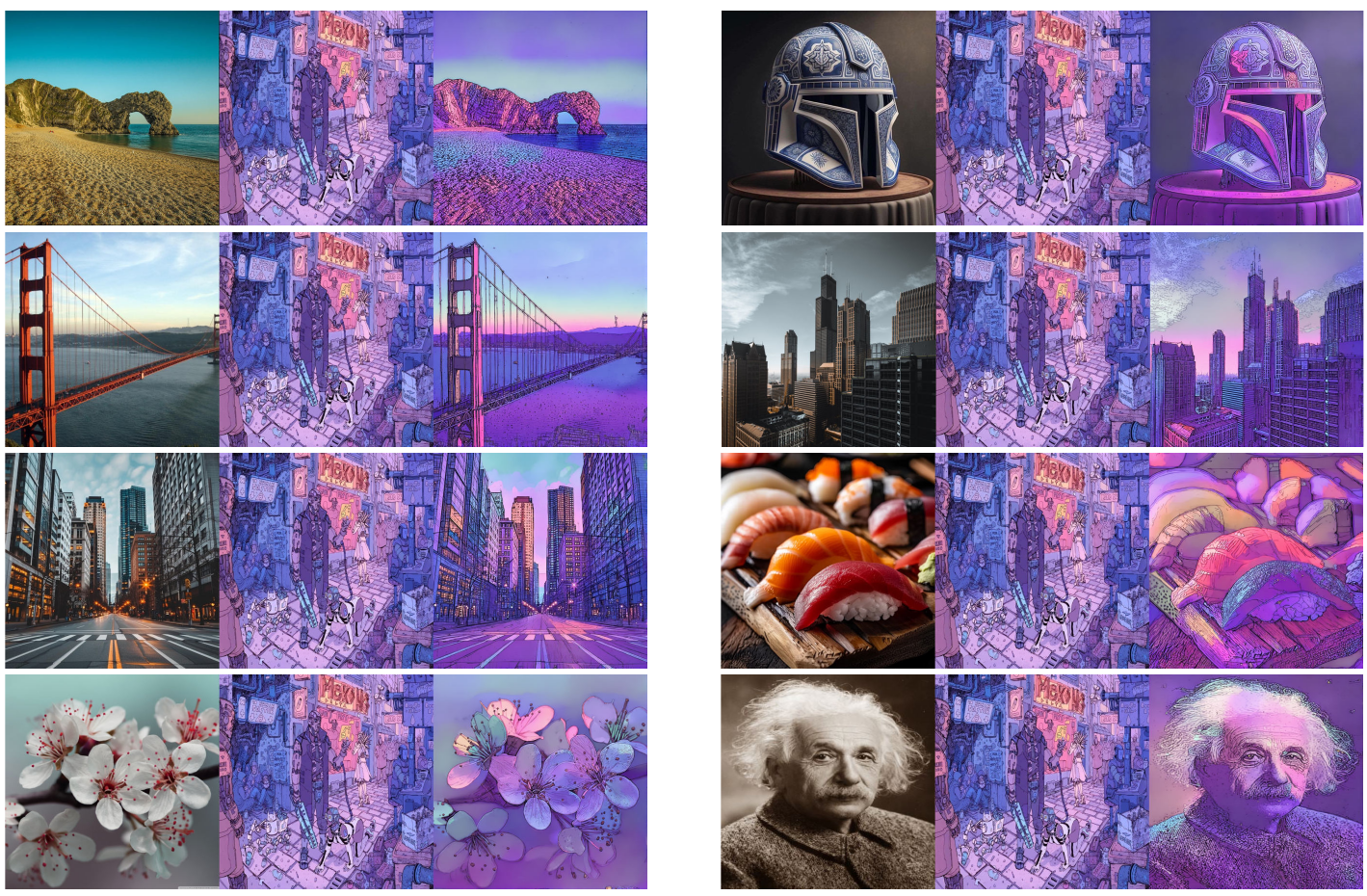}

    \caption{Additional image-guided style transfer results of OmniStyle. Left: content image, middle: style reference, right: stylized output.}
    \label{fig:res78}
\end{figure*}

\begin{figure*}
    \centering
    \includegraphics[width=1\textwidth]{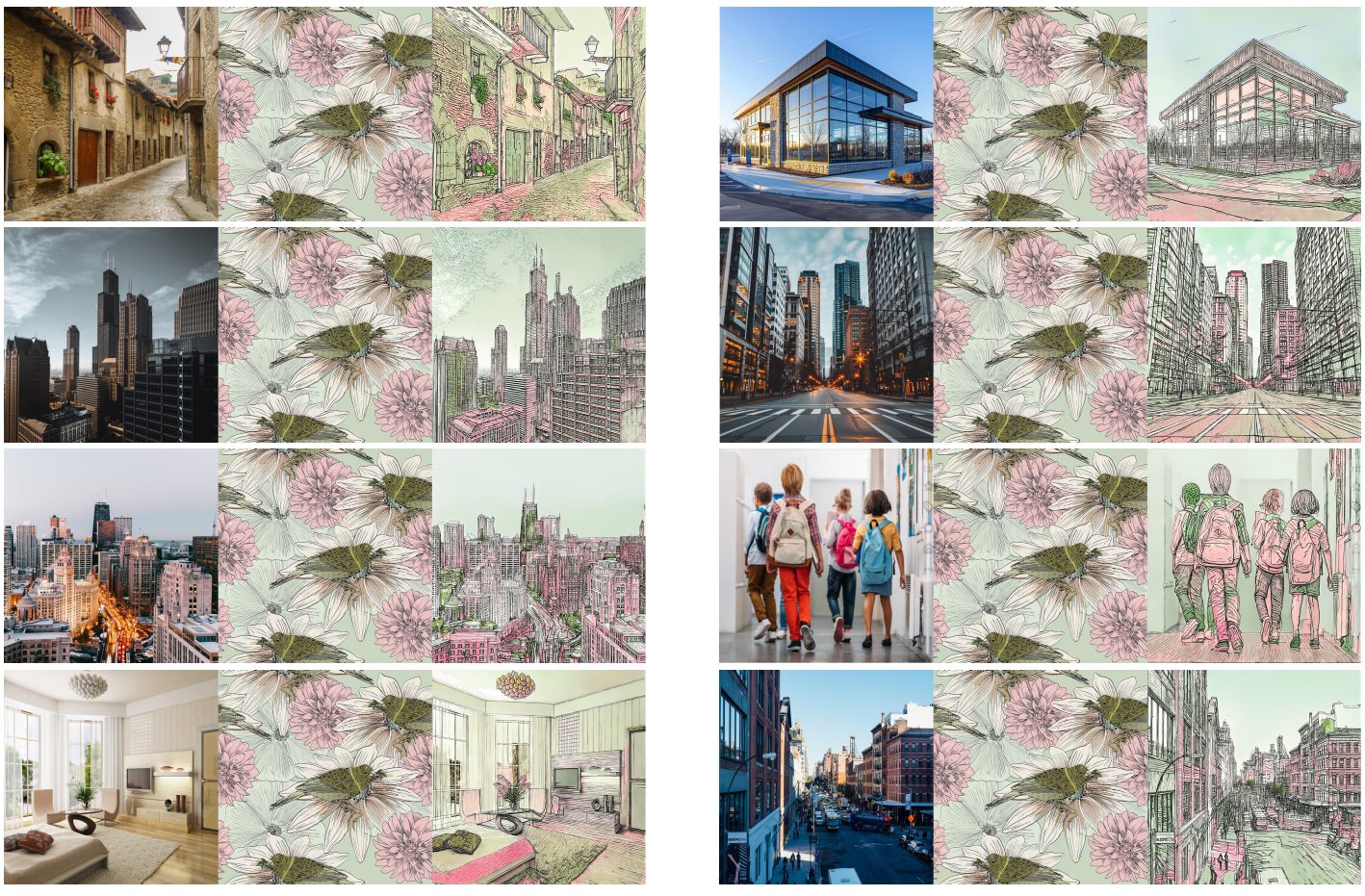}
    \includegraphics[width=1\textwidth]{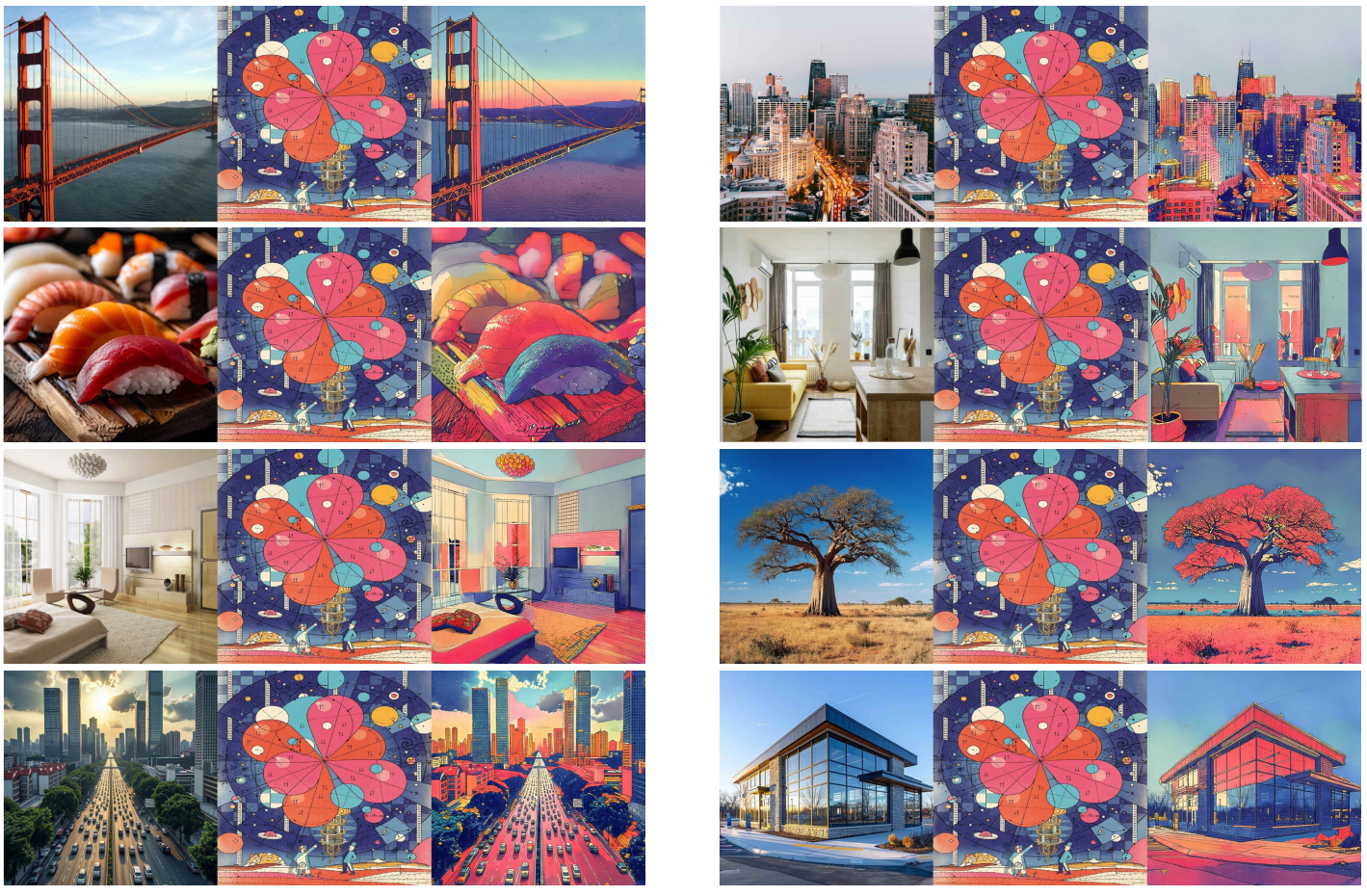}

    \caption{Additional image-guided style transfer results of OmniStyle. Left: content image, middle: style reference, right: stylized output.}
    \label{fig:res910}
\end{figure*}

\begin{figure*}
    \centering
    \includegraphics[width=1\textwidth]{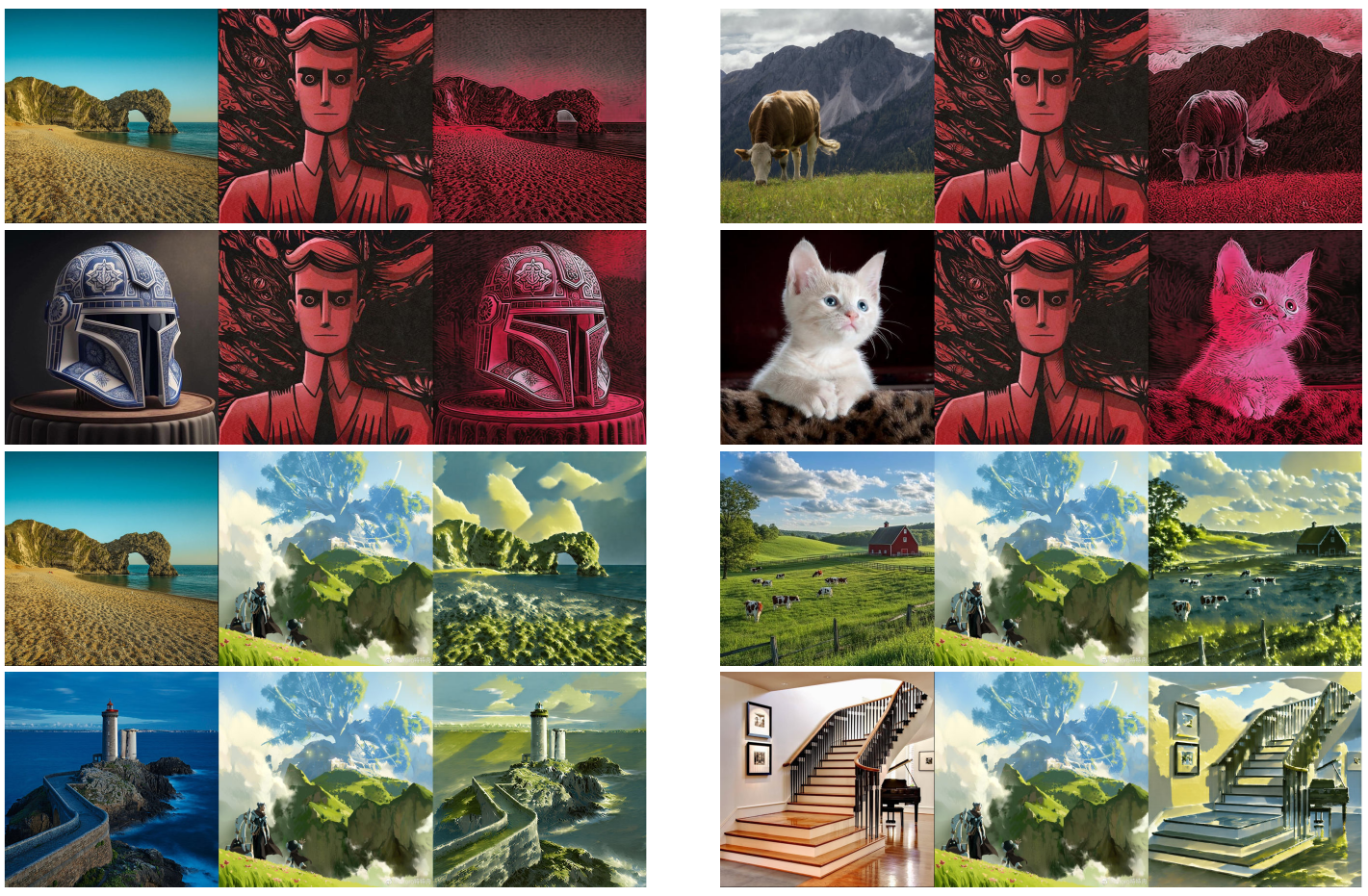}
    \includegraphics[width=1\textwidth]{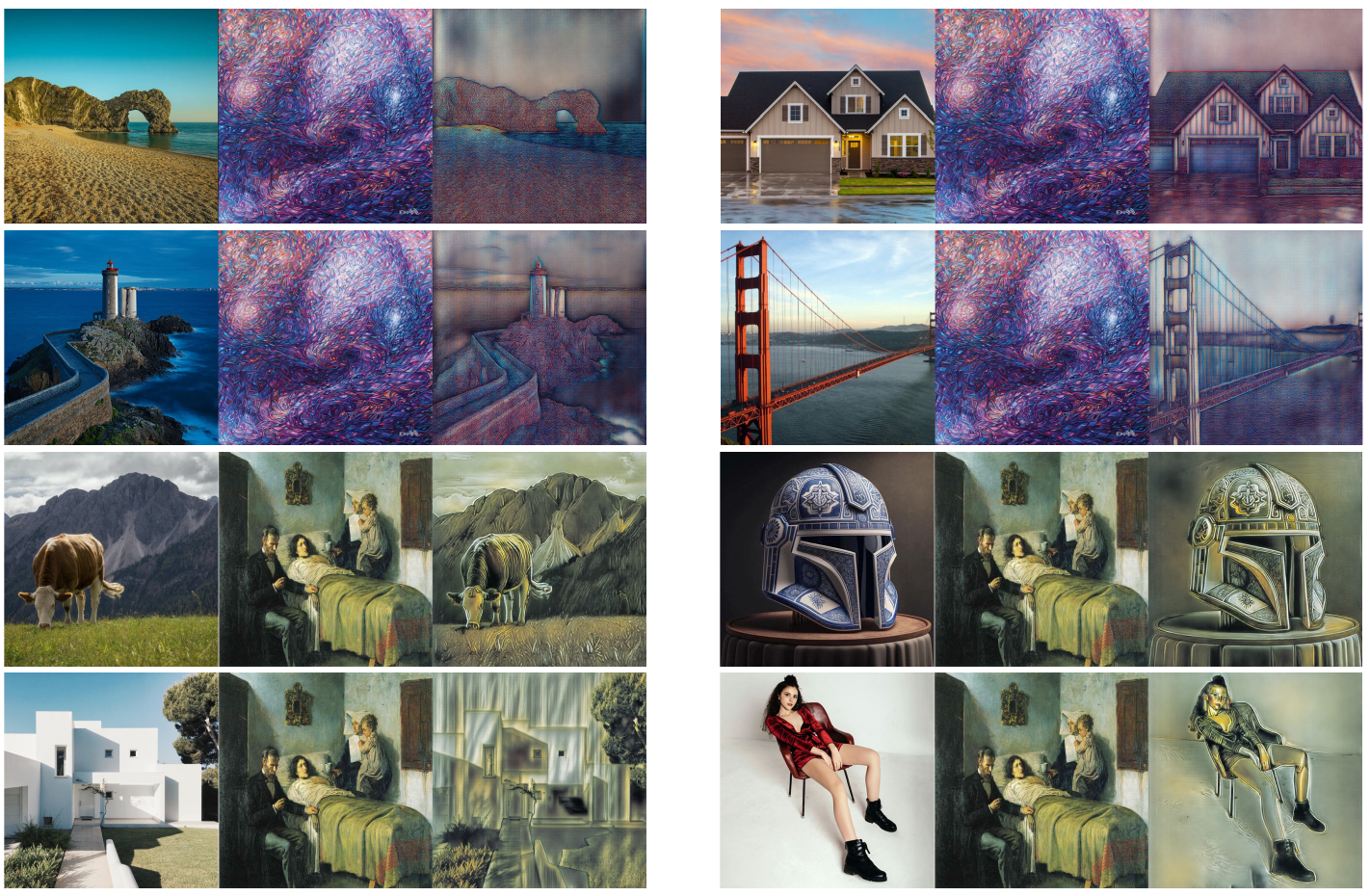}

    \caption{Additional instruction-guided style transfer results of OmniStyle. The left column shows the content images, the middle column presents the style references (Illustration Examples), and the right column displays the stylized outputs. Detailed instructions are omitted for brevity.}
    \label{fig:res1112}
\end{figure*}

\noindent{\textbf{Instruction-Based Style Transfer}}. As shown in Figures \ref{fig:res12}–\ref{fig:res910}, we present additional results of instruction-guided style transfer across various style categories. 
\noindent{\textbf{Image-Based Style Transfer}}. In Figures \ref{fig:res1112}, we provide more results of image-guided style transfer. 

These diverse style transfer outcomes demonstrate the superior performance of our method.

% \subsection{Further Applications}
% \noindent{\textbf{LoRA-Based Style Transfer.}}

\end{document}